\documentclass[10pt,twocolumn,letterpaper]{article}

\usepackage{cvpr}              % To produce the CAMERA-READY version
\usepackage{graphicx}
\usepackage{amsmath}
\usepackage{amssymb}
\usepackage{booktabs}
\usepackage{multirow}
\usepackage{tabularx}
\usepackage[ruled]{algorithm2e}
\usepackage[accsupp]{axessibility}  % Improves PDF readability for those with disabilities.
\usepackage{caption}
\usepackage{colortbl}
\usepackage{soul} % 导入 soul 包
\usepackage{color, xcolor} % 颜色包，color 必须导入，xcolor 建议导入
\definecolor{mygray}{gray}{0.95}
\usepackage{subcaption}
\usepackage[pagebackref,breaklinks,colorlinks]{hyperref}

\usepackage[capitalize]{cleveref}
\crefname{section}{Sec.}{Secs.}
\Crefname{section}{Section}{Sections}
\Crefname{table}{Table}{Tables}
\crefname{table}{Tab.}{Tabs.}
\begin{document}

%no pages in CVPR

\title{M$^{2}$-RAAP: A Multi-Modal Recipe for Advancing Adaptation-based Pre-training towards Effective and Efficient Zero-shot Video-text Retrieval}

\author{Xingning Dong\textsuperscript{\rm 1}, \quad Zipeng Feng\textsuperscript{\rm 1}, \quad Chunluan Zhou\textsuperscript{\rm 1}, 
\\ 
Xuzheng Yu\textsuperscript{\rm 1}, \quad Ming Yang\textsuperscript{\rm 1}, \quad Qingpei Guo\textsuperscript{\rm 1}{\footnotemark[2]}
\\
\normalsize{\textsuperscript{\rm 1}Ant Group}
\\
{\tt\small dongxingning1998@gmail.com, \quad fengzipeng.fzp@antgroup.com, \quad 	CZHOU002@e.ntu.edu.sg}
\\
{\tt\small 	yuxuzheng.yxz@antgroup.com, \quad m-yang4@u.northwestern.edu, \quad qingpei.gqp@antgroup.com}
}

% \maketitle
% \renewcommand{\thefootnote}{\fnsymbol{footnote}}
% \footnotetext[1]{Yutao Chen and Xingning Dong contribute equally to this manuscript.}
% \footnotetext[2]{Qingpei Guo is the Corresponding author.}

%%%%%%%%% ABSTRACT
\twocolumn[
{%
\maketitle
\renewcommand\twocolumn[1][]{#1}%
\begin{minipage}{0.98\linewidth}
         \vspace{0.0cm}
		 \includegraphics[width=1.0\textwidth]{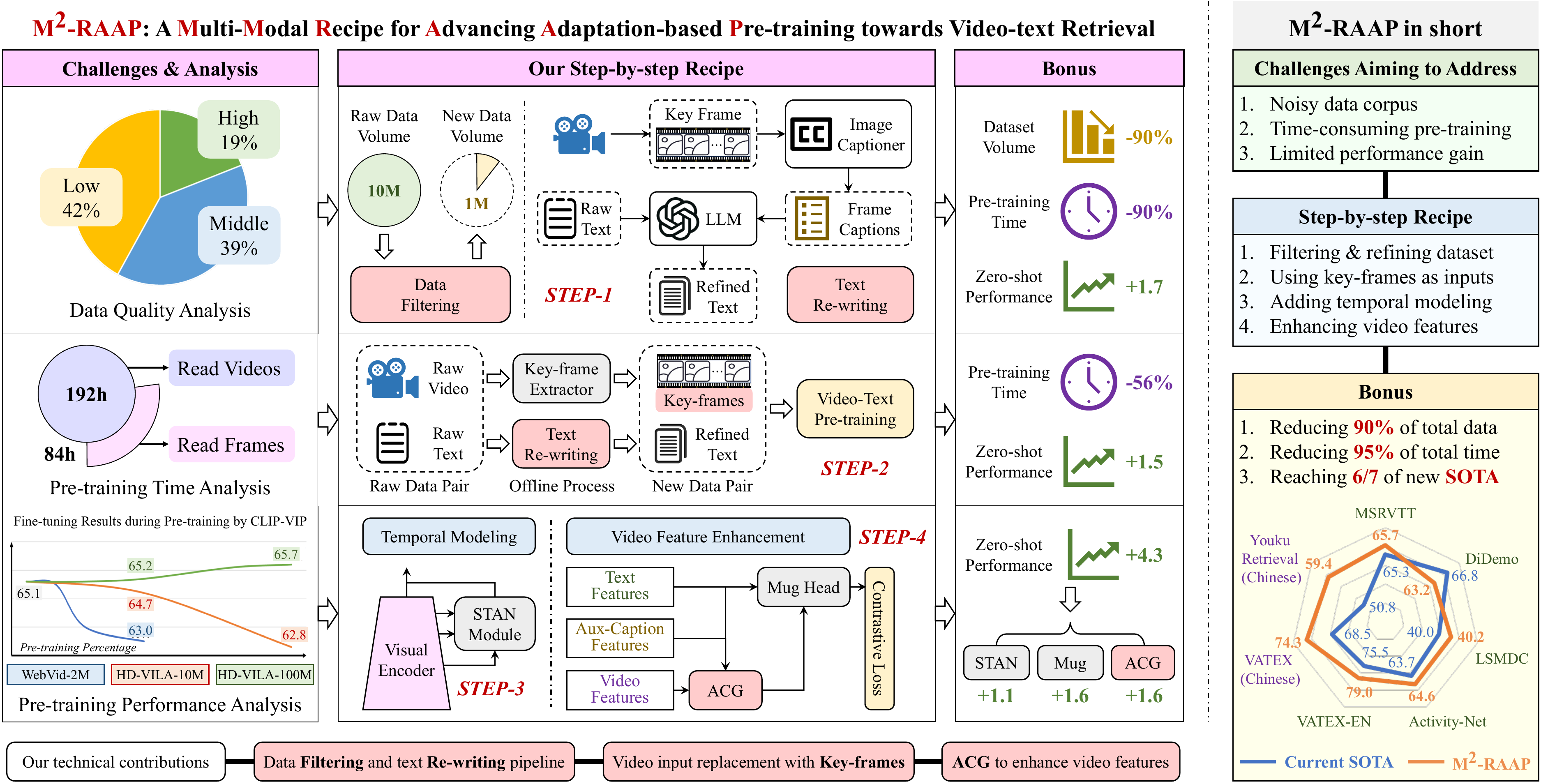}
		 \captionof{figure}{We propose M$^{2}$-RAAP, a multi-modal recipe for effective and efficient zero-shot video-text retrieval. Specifically, M$^{2}$-RAAP 1) filters and refines video-text pairs to improve the data quality, 2) adopts key-frames as video inputs to reduce pre-training time, and 3) introduces temporal modeling and video feature enhancement to promote pre-training performance. Compared with the baselines, M$^{2}$-RAAP employs only 10$\%$ of data volume (10M $\rightarrow$ 1M) and consumes only 5$\%$ of pre-training time (1920h $\rightarrow$ 92h), reaching a new SOTA on four English downstream zero-shot video-text retrieval datasets and two Chinese ones.}
		\label{fig:teaser}
		\vspace{0.8cm}
\end{minipage}
}]

\renewcommand{\thefootnote}{\fnsymbol{footnote}}
\footnotetext[2]{Qingpei Guo is the corresponding author.}

\begin{abstract}
   We present a \textbf{M}ulti-\textbf{M}odal \textbf{R}ecipe for \textbf{A}dvancing \textbf{A}daptation-based \textbf{P}re-training towards effective and efficient zero-shot video-text retrieval, dubbed \textbf{M$^{2}$-RAAP}. Upon popular image-text models like CLIP, most current adaptation-based video-text pre-training methods are confronted by three major issues, \textit{i}.\textit{e}., noisy data corpus, time-consuming pre-training, and limited performance gain. Towards this end, we conduct a comprehensive study including four critical steps in video-text pre-training. Specifically, we investigate 1) data filtering and refinement, 2) video input type selection, 3) temporal modeling, and 4) video feature enhancement. We then summarize this empirical study into the M$^{2}$-RAAP recipe, where our technical contributions lie in 1) the data filtering and text re-writing pipeline resulting in 1M high-quality bilingual video-text pairs, 2) the replacement of video inputs with key-frames to accelerate pre-training, and 3) the Auxiliary-Caption-Guided (ACG) strategy to enhance video features. We conduct extensive experiments by adapting three image-text foundation models on two refined video-text datasets from different languages, validating the robustness and reproducibility of M$^{2}$-RAAP for adaptation-based pre-training. Results demonstrate that M$^{2}$-RAAP yields superior performance with significantly reduced data (-90$\%$) and time consumption (-95$\%$), establishing a new SOTA on four English zero-shot retrieval datasets and two Chinese ones. We are preparing our refined bilingual data annotations and codebase, which will be available at \href{https://github.com/alipay/Ant-Multi-Modal-Framework/tree/main/prj/M2_RAAP}{https://github.com/alipay/Ant-Multi-Modal-Framework/tree/main/prj/M2$\_$RAAP}.
   % We will release refined bilingual data annotations and codebase to facilitate future research.

\end{abstract}

%%%%%%%%%%%%%%% Introduction %%%%%%%%%%%%%%%

\section{Introduction}
\label{sec:introduction}

Pre-trained foundation models fine-tuned to downstream tasks have achieved remarkable progress in both NLP \cite{sanh2019distilbert, joshi2020spanbert, devlin2018bert} and CV fields \cite{zhu2020actbert, huang2020pixel, lu2019vilbert}, inspiring extensive research efforts to extend this ``Pre-training $\&$ Fine-tuning'' paradigm to the video-text field \cite{xu2021vlm, li2020hero, sun2019videobert, dong2023snp}. Video-text pre-training targets to derive a universal model capable of adapting to various downstream cross-modal tasks \cite{yang2021just, sun2021video, patel2021recent, nie2022search}. Among these tasks, video-text retrieval \cite{luo2021clip4clip, bain2021frozen, ge2022bridging, jiang2023dual} emerges as a pivotal function, owing to its potential for seamless applications to short video apps for search and recommendation. In this paper, we tackle video-text retrieval in a zero-shot setting, which requires models to infer on the testing set without additional fine-tuning to fit an unseen data distribution, and thoroughly evaluate the generalization capability of video-text models.

To develop a robust video-text pre-trained model, current methods mainly pre-train a plain model directly from scratch \cite{lei2021less, bain2021frozen, luo2020univl, yang2021taco} or adapt a well-trained image-text model \cite{luo2021clip4clip, fang2021clip2video, xue2022clip, liu2023mug}. Compared with the training-from-scratch scheme, the adaptation-based paradigm may take full advantage of well-learned knowledge embedded in image-text foundation models, which usually yields better performance and consumes less pre-training time than the training-from-scratch one. Therefore, we follow the line of adaptation-based approaches and strive to empower an existing image-text model like CLIP \cite{radford2021learning} to facilitate video-text retrieval. 

After intensive research, adaptation-based methods have not realized their potential to push significant performance gain given a robust image-text model. Upon this observation, we thoroughly investigate the issues that may restrict adaptation-based video-text pre-training and summarize them into three primary challenges, as shown in the left part of Figure \ref{fig:teaser}: 1) \textbf{Noisy data corpus}. Mug-STAN \cite{liu2023mug} has quantitatively assessed video-text misalignment in the widely-employed WebVid-2.5m dataset \cite{bain2021frozen}, by employing CLIP to compute frame-text similarity scores. They define videos with more than 2/3 frames whose scores exceed 0.5 as high-quality, while videos with less than 1/3 frames are deemed to be low. Under this criteria, only 1/5 of videos exhibit commendable consistency. Conversely, 2/5 of videos are noisy and may distract model optimization. 2) \textbf{Time-consuming pre-training}. Current methods typically require a quite long time for pre-training. \textit{E}.\textit{g}., UMT-L \cite{li2023unmasked} needs 130 hours (5.4 days) for pre-training on 25M data pairs with 32 A100 GPUs. Beyond compressing the data volume, we figure out that replacing raw video inputs with offline extracted key-frames would considerably accelerate the pre-training process, halving the total time from 192h to 84h (-56$\%$).
% We evaluate the time consumption of whether feed models with raw videos or offline extracted frames under the same experimental setting. The results demonstrate that directly reading frames rather than raw videos would significantly accelerate the pre-training procedure, reducing the overall time from 192h to 84h (-56$\%$). 
3) \textbf{Limited performance gain}. This issue has been raised by CLIP-VIP \cite{xue2022clip} and Mug-STAN \cite{liu2023mug}, where persistently adapting image-text models on current video-text datasets results in negligible gain and even leads to a decline in performance. Specifically, CLIP-VIP reports the metric of the average value of Recall@1/5/10 on MSRVTT (dubbed \textit{MSRVTT-AVG-R}). Results show the performance only outperforms the baseline (65.1) by +0.6 after pre-training on HD-VILA-100M (65.7), while exhibiting a noticeable decrease on WebVid-2.5M (63.0) by -2.1 and HD-VILA-10M (62.8) by -2.3 at \textit{MSRVTT-AVG-R}.

The above challenges indicate three promising directions to improve the adaptation-based video-text pre-training. However, there lacks a comprehensive study on the expected performance and efficiency gain brought by addressing these issues. Recently, VINDLU \cite{cheng2023vindlu} has carried out an empirical study demystifying six essential components and their contributions to a robust training-from-scratch scheme, providing valuable insights to future research. Inspired by VINDLU, we aim to fill in the gap of this empirical study in adaptation-based video-text pre-training, resulting in the proposed M$^{2}$-RAAP recipe.

\begin{figure}[t]
	\centering
    	\begin{subfigure}{1\linewidth}
    		\includegraphics[width=1.0\textwidth]{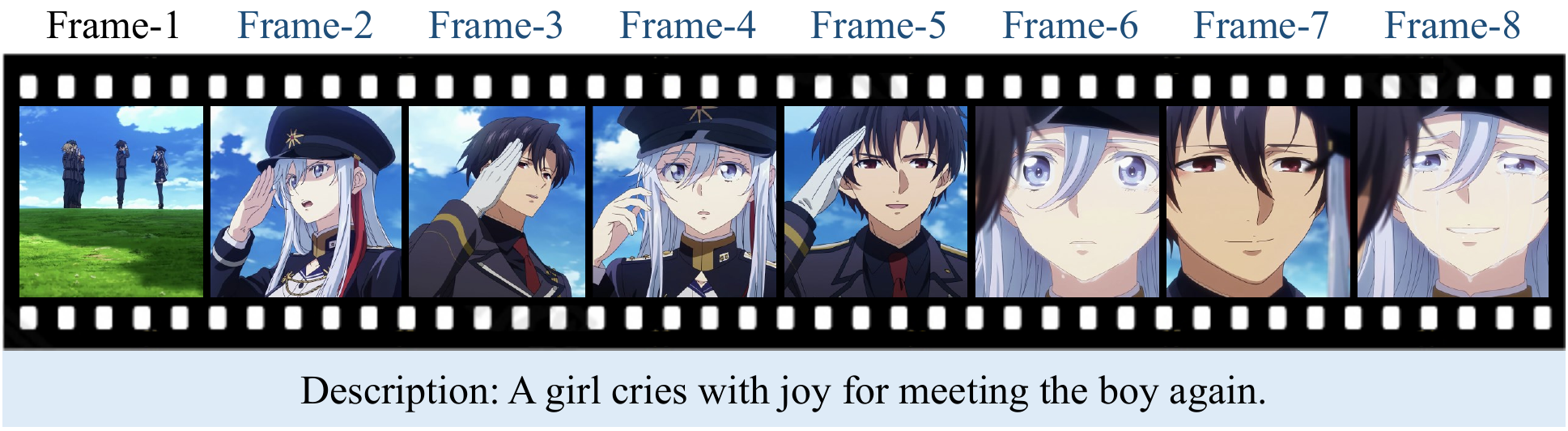}
    		\caption{\scriptsize{Though there is only one person in most of the frames (Frames 2-8), one can still understand this video describes a ``girl meets boy'' story by exploiting temporal cues.}}
    		\label{fig:intro-1}
    	\end{subfigure}

            \vspace{0.2cm}
     
    	\begin{subfigure}{1\linewidth}
    		\includegraphics[width=1.0\textwidth]{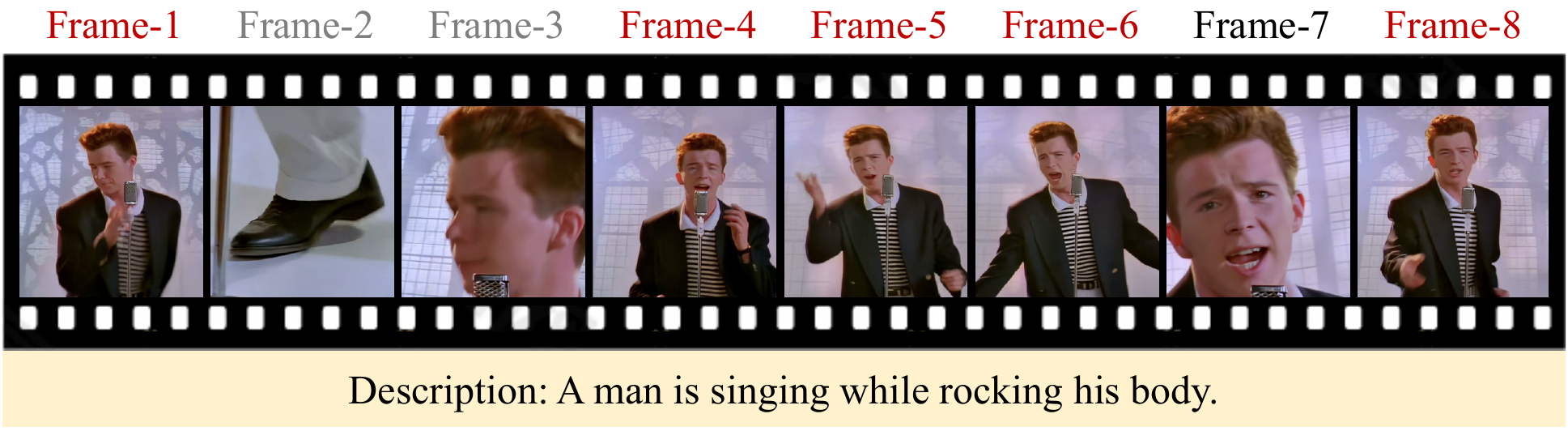}
    		\caption{\scriptsize{Compared with Frames 2-3, the frames marked in red are more critical to understand the action of ``rocking'', whose features should be enhanced.}}
    		\label{fig:intro-2}
    	\end{subfigure}
    	
	\vspace{-0.2cm}
	\caption{Two examples that demonstrate the importance of temporal modeling and video feature enhancement.}
	\vspace{-0.4cm}
	\label{fig:intro-all}
\end{figure}

M$^{2}$-RAAP is a step-by-step recipe that seeks to answer the question, ``What are the key steps and how do they contribute to an effective and efficient adaptation-based video-text pre-training method". M$^{2}$-RAAP begins with a robust image-text model (\textit{e}.\textit{g}., CLIP) and employs a simple progressive expansion scheme. During each step, we integrate additional modules or strategies, as illustrated in the middle part of Figure \ref{fig:teaser}. We then quantatively evaluate the performance and efficiency gain that are attributable to the previous implemented operation, as depicted in the right part of Figure \ref{fig:teaser}. 

M$^{2}$-RAAP first develops an automatic and cost-efficient data filtering and text rewriting pipeline, addressing the critical demand for a high-quality data corpus in video-text pre-training, while yielding 1M high-quality bilingual video-text pairs. By leveraging advanced large image captioners \cite{li2023blip} and large language models (LLM) \cite{achiam2023gpt}, we filter misaligned data pairs and promote the quality of text annotations. In this way, we reduce 90$\%$ of pre-training data (10M $\rightarrow$ 1M) and time consumption, while obtaining a notable performance gain of +1.7 at \textit{MSRVTT-AVG-R}. 

M$^{2}$-RAAP then advocates for using key-frames as a cost-efficient alternative to raw videos for pre-training. Key-frames are selected to promote the diversity among extracted frames, which intends to choose those informative ones to convey the video content. By employing offline key-frames to skip the time-consuming video decoding operation, M$^{2}$-RAAP reduces 56\% of training time while obtaining a performance gain of +1.5 at \textit{MSRVTT-AVG-R}.

More importantly, M$^{2}$-RAAP aims to further push for performance gain of adaptation-based pre-training by leveraging the intrinsic properties of videos, 
encouraging the integration of temporal modeling mechanisms and video feature enhancement strategies. 

Videos are beyond sequences of 2D images along the temporal axis, whose semantic contents are largely conveyed by the interaction among critical frames. As shown in Figure \ref{fig:intro-1}, even most of the frames depict a single person, this video describes a “girl meets boy” story after harnessing temporal cues of all frames. 
Figure \ref{fig:intro-2} illustrates that not all frames contribute equally to understanding the caption, \textit{i}.\textit{e}., Frames 2-3 are misaligned with the action ``rocking''. These two examples indicate that video understanding requires enhancing video features by advanced temporal modeling and down-weighing inconsistent frames. 

In this paper, besides employing Mug-STAN \cite{liu2023mug}, we propose a novel Auxiliary-Caption-Guided (ACG) strategy to leverage auxiliary frame captions for explicit video feature enhancement, yielding a performance gain of +1.6 at \textit{MSRVTT-AVG-R}. Specifically, ACG incorporates an additional image-text contrastive learning towards frame-caption pairs, thereby preserving well-learned knowledge embedded in image-text models. Moreover, ACG dynamically re-weights the contributions of frame features from two aspects: the inter-modal similarity between captions and frames, and the intra-modal one within captions and the given text.

To validate that M$^{2}$-RAAP is a reliable and reproducible recipe, we implement M$^{2}$-RAAP with \textbf{three} basic image-text models on \textbf{two} pre-training datasets, and evaluate M$^{2}$-RAAP on \textbf{seven} downstream zero-shot retrieval testing sets in both English and Chinese. M$^{2}$-RAAP reduces 90$\%$ of pre-training data (10M $\rightarrow$ 1M) and 95$\%$ of time cost (1920h $\rightarrow$ 92h), reaching a new SOTA on six zero-shot testing sets with a total of 11.5 hours pre-trained on 8 A100 GPUs.

In summary, we propose M$^{2}$-RAAP, a step-by-step recipe to promote adaptation-based pre-training with the following four technical 
contributions: 

\vspace{-0.1cm}

\begin{itemize}
\setlength{\itemsep}{0pt}
\setlength{\parsep}{0pt}
\setlength{\parskip}{0pt}

	% \item[1.] We propose M$^{2}$-RAAP, a step-by-step recipe to effectively and efficiently promote zero-shot video-text retrieval performance.
	
	\item We develop an automatic and cost-efficient data filtering and text rewriting pipeline, resulting in 1M high-quality bilingual video-text pairs. In this way, we reduce 90$\%$ of data volume, yielding a performance gain of +1.7 at \textit{MSRVTT-AVG-R}.
	
	\item We demonstrate that key-frames are superior and more cost-efficient inputs for pre-training. In this way, we reduce 56$\%$ of pre-training time, and yielding a performance gain of +1.5 at \textit{MSRVTT-AVG-R}.
	
	\item We propose ACG, a novel strategy to enhance video features, yielding a performance gain of +1.6 at \textit{MSRVTT-AVG-R}.

    \item We conduct extensive experiments and ablation study on three basic image-text models, two pre-training datasets, and seven downstream retrieval testing sets in English and Chinese. Results affirm the robustness of M$^{2}$-RAAP as a sound multi-modal recipe for effective and efficient video-text retrieval.

\end{itemize}

%%%%%%%%%%%%%%% Related Work %%%%%%%%%%%%%%%

\section{Related Work}

\noindent\textbf{Image-text Pre-training.}

Recent years have witnessed remarkable progress in image-text pre-training methods \cite{chen2022altclip, lu2019vilbert, yu2022coca}, where CLIP \cite{radford2021learning} is one of the most renowned models widely applied in various downstream image understanding tasks \cite{qu2020context, cheng2023sample, han2023semantic}. To thoroughly assess the contributions of different components within the image-text model, METER \cite{dou2022empirical} has conducted a detailed empirical study, providing valuable insights for future research in image-text pre-training.

% ---

\begin{figure*}[t]
	\begin{center}
		\includegraphics[width=1\textwidth]{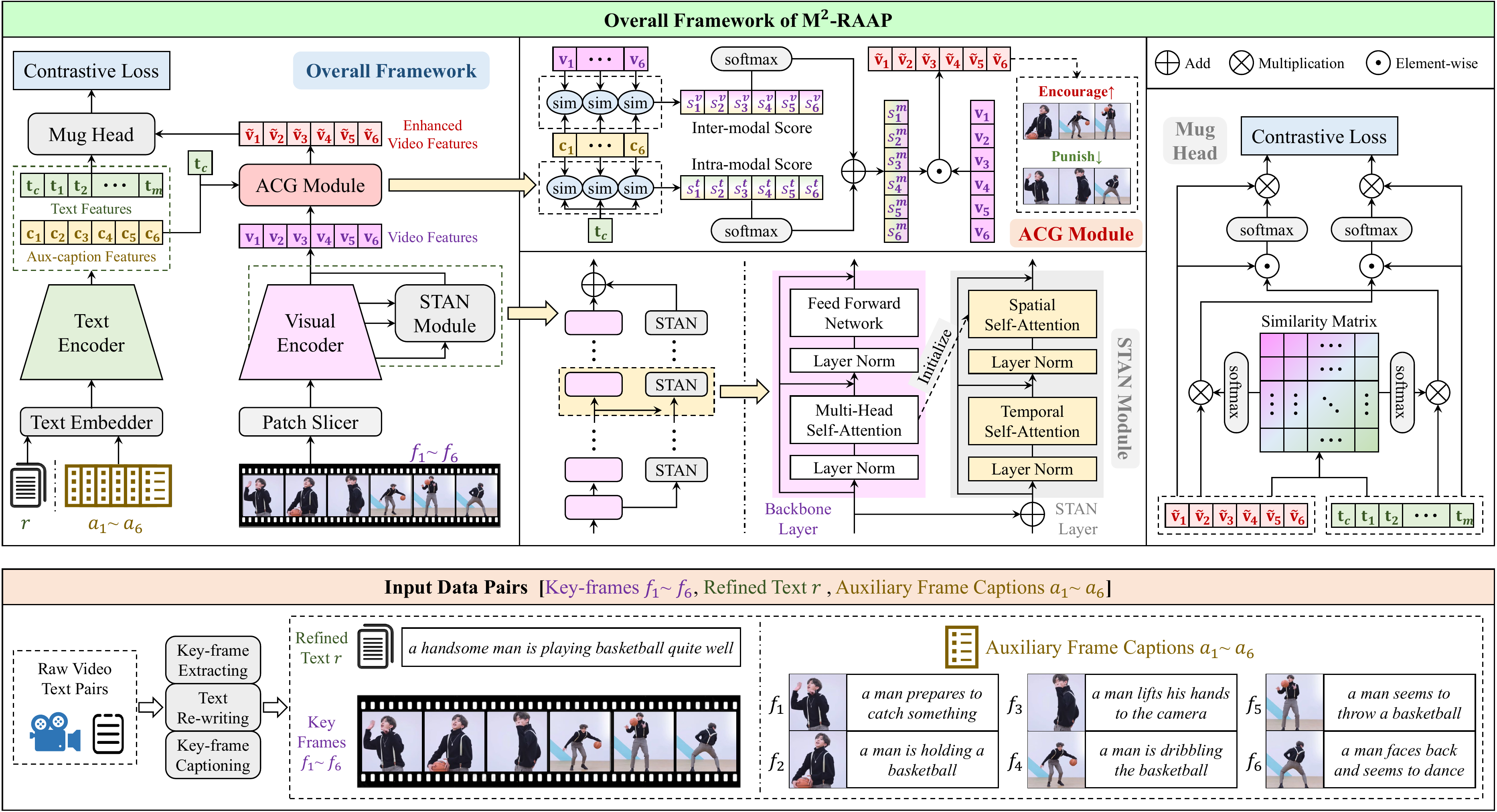}
	\end{center}
	\vspace{-0.4cm}
	\caption{The pipeline of M$^{2}$-RAAP. M$^{2}$-RAAP employs a progressive expansion scheme to evaluate the contributions of each component. We illustrate the architectures of the overall pre-training framework (top-left part), STAN module (middle-center part), Mug head (top-right part), and our proposed ACG strategy (top-center part).}
	
	\vspace{-0.2cm}
	\label{fig:framework}
\end{figure*}

\noindent\textbf{Large-scale Video-text Datasets.}

Large-scale high-quality datasets are a prerequisite for the successful applications of the ``Pre-training $\&$ Fine-tuning'' paradigm. Various image-text foundation models are pre-trained on large-scale image-text datasets like COCO \cite{lin2014microsoft}, CC \cite{sharma2018conceptual}, and LAION \cite{schuhmann2022laion}. However, compared with image-text pairs that are easily to collect and usually have good quality in practice, video-text pairs are generally more noisy with a much higher cost to refine their correspondences. \textit{E}.\textit{g}., HowTo100M \cite{miech2019howto100m}, CNVid-3.5M \cite{gan2023cnvid}, and Youku-mPLUG-10M \cite{xu2023youku} are all collected from English or Chinese websites and employ the corresponding ASR text or subtitles as annotations, encountering a severe video-text inconsistency problem. Meanwhile, for another widely-employed WebVid-10M \cite{bain2021frozen} dataset, their videos are partially misaligned with the corresponding text, as described in the Sec. \ref{sec:introduction}. Therefore, data quality becomes one of the primary obstacles in video-text pre-training. Towards this end, thanks to powerful large image captioners like BLIP2 \cite{li2023blip} and large language models like GPT-4 \cite{achiam2023gpt}, we develop an automatic and cost-efficient data filtering and text re-writing pipeline to improve the quality of the video-text corpus, thereby facilitating both English and Chinese video-text pre-training.

% ---

\noindent\textbf{Video-text Pre-training.} 

Video-text pre-training is a promising direction and attracts increasing interest in the field of multi-modal learning, which could be roughly divided into two categories: 1) pre-training from scratch \cite{lei2021less, bain2021frozen, xu2021vlm, luo2020univl}, and 2) adapting robust image-text models \cite{luo2021clip4clip, fang2021clip2video, li2023unmasked, liu2023mug}. In contrast to the training-from-scratch methods that start with nearly uninitialized models, adaptation-based methods leverage well-learned knowledge embedded in robust image-text models, which usually achieve fast convergence and superior performance. With the rapid development in video-text pre-training, current methods are evolving to exhibit increasing complexity in model designs and pre-training protocols. Therefore, there requires a comprehensive study like METER \cite{dou2022empirical} to decipher which factors are critical for video-text pre-training. Upon this observation, VINDLU \cite{cheng2023vindlu} carries out a thorough investigation covering a broad range of factors among various training-from-scratch methods, based on which they summarize a step-by-step recipe to facilitate future research. 

Nevertheless, there still lacks an comprehensive study like METER \cite{dou2022empirical} and VINDLU \cite{cheng2023vindlu} in adaptation-based video-text pre-training, resulting in under-exploration of the quantitative contributions of components in the adaptation-based scheme. 
% Simultaneously, we notice that several methods point out different challenges towards adaptation-based video-text pre-training. \textit{E}.\textit{g}., CLIP4Clip aims to address the fusion problem in transferring image-text models, Mug-STAN \cite{liu2023revisiting} highlights that non-generalizable temporal modeling and partially misaligned video-text data would impede the model adaptation, and CLIP-VIP \cite{xue2022clip} discovers an over-fitting phenomenon that video-text pre-training only yields limited performance gain. 
After a thorough investigation of pioneering methods \cite{xue2022clip, luo2021clip4clip, liu2023mug, liu2023revisiting}, we propose M$^{2}$-RAAP to fill in this gap, aiming to gain an overall insight to reveal which factors yield more significance to an effective and efficient adaptation-based video-text pre-training.

%%%%%%%%%%%%%%% Methodology %%%%%%%%%%%%%%%

\begin{figure*}[t]
	\begin{center}
		\includegraphics[width=1\textwidth]{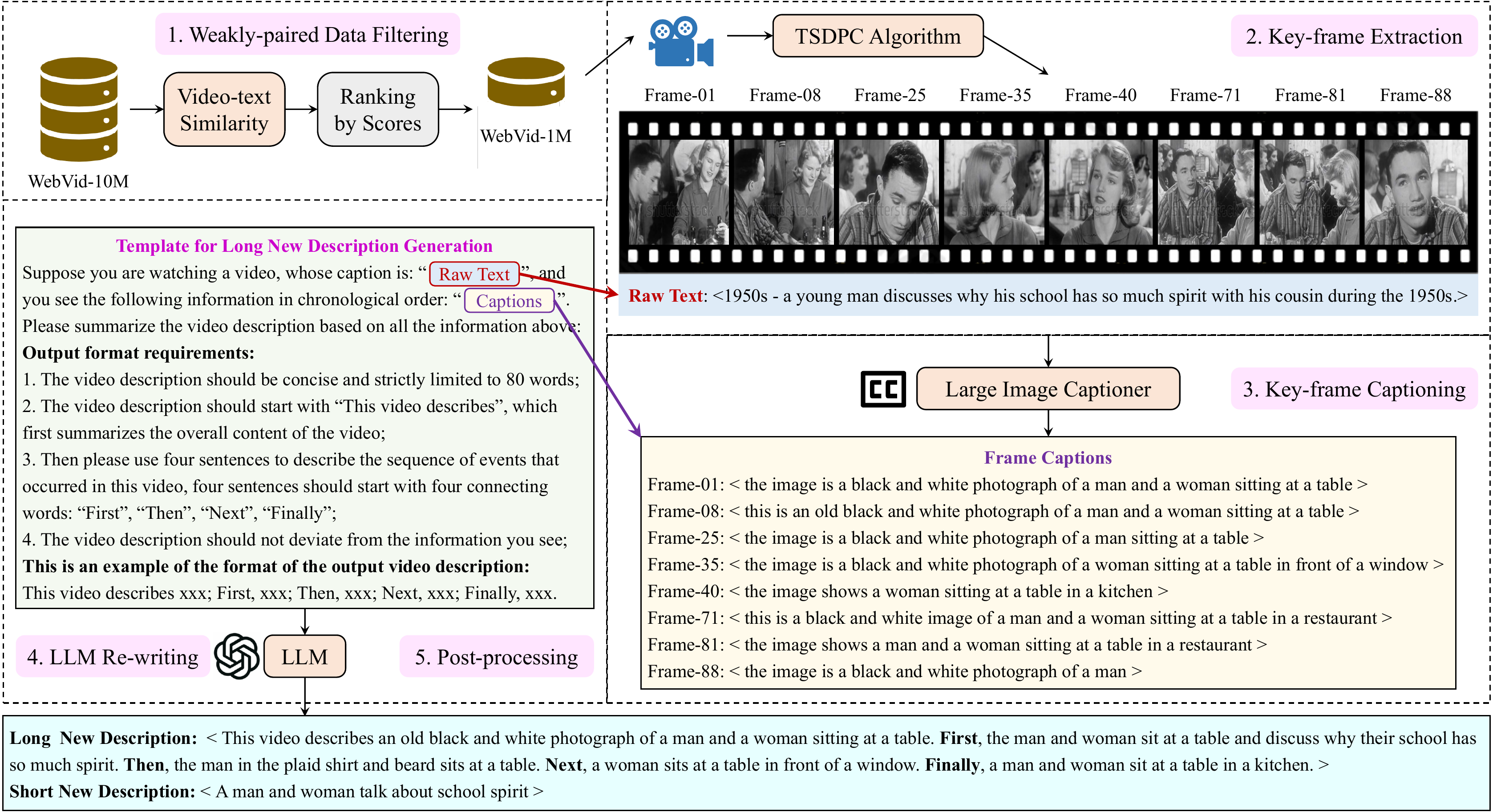}
	\end{center}
	\vspace{-0.4cm}
	\caption{An example of the automatic data filtering and text re-writing pipeline on the English WebVid-10M dataset.}
	\vspace{0.2cm}
	\label{fig:dataset}
\end{figure*}

\begin{figure*}[t]
	\begin{center}
		\includegraphics[width=1\textwidth]{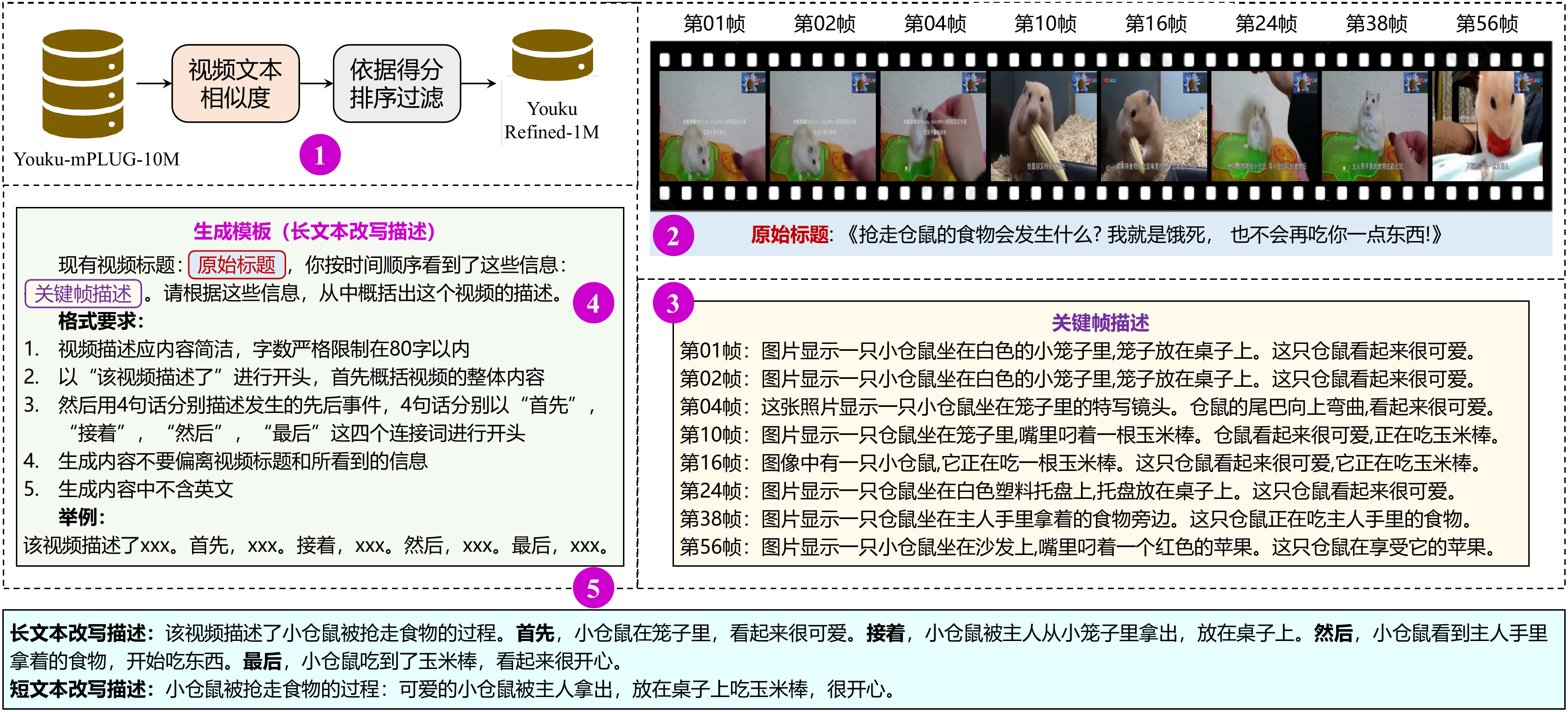}
	\end{center}
	\vspace{-0.4cm}
	\caption{An example of the automatic data filtering and text re-writing pipeline on the Chinese Youku-mPLUG-10M dataset.}
	
	\vspace{-0.2cm}
	\label{fig:dataset-cn}
\end{figure*}

\section{Methodology of M$^{2}$-RAAP Recipe}
\label{method:all}

In this section, we elaborate on our M$^{2}$-RAAP recipe for effective and efficient zero-shot video-text retrieval through adaptation-based pre-training. We start with a conventional image-text model, and progressively extend it to a final one, where the overall framework is illustrated in the top-left part of Figure \ref{fig:framework}. At each step, M$^{2}$-RAAP attempts to quantitatively analyze the performance and efficiency gain brought by addressing primary issues including data quality, video input type selection, and model design.

% -----------------------------------

\subsection{Step 0: Beginning Ingredients}

We start with a simple baseline that directly adapts a widely-employed image-text model CLIP with a renowned video-text dataset WebVid-10M. CLIP is a typical twin-tower-based model containing a visual encoder ${E}_{vis}$ and a text encoder ${E}_{txt}$.

Given a mini-batch (denoted as $\mathcal{B}$) of videos $\{{v}^{b}\}_{b=1}^\mathcal{B}$ and their corresponding refined text annotations $\{{r}^{b}\}_{b=1}^\mathcal{B}$, we first sparsely (and randomly) sample ${N}_{f}$ frames from each video. (${N}_{f}$ is usually much smaller than the total number of frames in this video). We then slice each frame into patches, obtaining visual patch embeddings. Simultaneously, a frozen tokenizer is employed to process the input text $r$ into fixed-length text embeddings $\mathbf{W} = [{\mathbf{w}}_{cls}, {\mathbf{w}}_{1}, {\mathbf{w}}_{2}, \cdots, {\mathbf{w}}_{{N}_{t}-1}]$, $\mathbf{W} \in {\mathbb{R}}^{{N}_{t}*d}$, where ${N}_{t}$ is the length of text tokens and $d$ is the dimension of the embedding.

Afterward, we feed the visual encoder ${E}_{vis}$ with patch embeddings, obtaining the [\textit{CLS}] tokens of all frame patches as video features $\mathbf{V} \in {\mathbb{R}}^{{N}_{f}*d}$. Meanwhile, we utilize the text encoder ${E}_{txt}$ to process text embeddings $\mathbf{W}$, obtaining text features $\mathbf{T} \in {\mathbb{R}}^{{N}_{t}*d}$.

Ultimately, we utilize the video-text-contrastive (VTC) proxy task as the optimization objective of video-text pre-training. VTC aims to promote cross-modal alignment by maximizing the similarity matrix of video-text pairs, which could be formulated as: 
\begin{equation}
\label{eq:VTC}
	\mathcal{L}_{vtc}(\overline{\mathbf{v}}, \mathbf{t}) = -\sum_{b=1}^{\mathcal{B}} {\rm log}   \frac{{\rm exp}^{ \langle \overline{\mathbf{v}}^{b}, \mathbf{t}^{b}_{cls} \rangle} }
	{ {\rm exp}^{ \langle \overline{\mathbf{v}}^{b}, \mathbf{t}^{b}_{cls} \rangle} + \sum_{e \neq b}{\rm exp}^{ \langle \overline{\mathbf{v}}^{e}, \mathbf{t}^{b}_{cls} \rangle} },
\end{equation}
where $\overline{\mathbf{v}}$ represents mean-pooling products of video features $\sum_{i=1}^{{N}_{f}}(\mathbf{v}_{i})$, $\mathbf{t}_{cls}$ denotes the [\textit{CLS}] token of text features $\mathbf{T}$, and $\langle \cdot , \cdot  \rangle$ denotes the matrix multiplication operation.

As aforementioned, current adaptation-based video-text pre-training suffers from three primary obstacles, \textit{i}.\textit{e}., 1) noisy data corpus, 2) time-consuming pre-training, and 3) limited performance gain. In the next several subsections, we progressively expand this simple baseline to address these three challenges.

% -----------------------------------

\subsection{Step 1: Data Filtering and Refinement}
\label{method:step-1}

We first develop an automatic and cost-efficient data filtering and text rewriting pipeline, which improves the data quality for video-text pre-training. Different from images, collecting a large-scale high-quality video dataset is usually intractable, since only partially aligned text annotations like ASR and subtitles are available. Moreover, hiring skilled human annotators to label millions of videos is extremely expensive and practically infeasible. Therefore, there is a critical demand for the construction of a high-quality corpus for video-text pre-training, with little progress so far in the field. Fortunately, remarkable progress has emerged in large image captioners and large language models (LLM), based on which we design an effective and cost-efficient text refinement pipeline without extensive human intervention.

We start with two widely-employed datasets: WebVid-10M \cite{bain2021frozen} in English and Youku-mPLUG-10M \cite{xu2023youku} in Chinese. As illustrated in Figures \ref{fig:dataset} (English) and \ref{fig:dataset-cn} (Chinese), the overall pipeline contains the following five key steps.

1) \textbf{Weakly-paired data filtering}. As pointed out by CNVid \cite{gan2023cnvid}, those noisy and misaligned video-text pairs would hinder the video-text pre-training. Therefore, we follow the instructions of CNVid and employ robust bilingual image-text models to calculate similarity scores between videos and their associated text annotations. We then sort consistency scores and retain the top 1M video-text pairs by their ranks. 

2) \textbf{Key-frame extraction}. Key-frames are crucial as they could highly generalize the video content. We exploit an unsupervised key-frame extraction algorithm, named temporal segment density peaks clustering (TSDPC) \cite{tang2023deep}, to obtain 8 frames for each video. 

3) \textbf{Key-frame captioning}. We then employ a large bilingual image captioner to generate detailed descriptions for each extracted key-frame. We name these generated text as auxiliary captions.

4) \textbf{LLM re-writing}. Afterward, we organize raw text annotations and 8 associated auxiliary captions $\{{a}_{i}\}_{i=1}^{8}$ into a pre-defined template. We then instruct LLM to pretend to watch the given video through the input prompt, and require LLM to output a new description that highly generalizes the whole event within the video. In this step, we employ two templates to derive both detailed long descriptions ($\approx$ 80 words) and simplified short ones ($\leq$ 15 words) from LLM, which benefits both long-text and short-text video retrieval tasks. Specifically, we expect the generated long descriptions to contain rich temporal information of events that happened in videos, while requiring short descriptions to summarize the video content as simply as possible.

5) \textbf{Post-processing}. Though current LLM presents superior capability in text processing, their outputs are sometimes nonsense due to the inherent hallucination problem. Besides, it is essential to trim the format of newly re-written annotations. Therefore, we implement a post-processing scheme to filter those generated results that are too short, too long, or nearly meaningless. 

Following this pipeline, we obtain 1M English video-text pairs from WebVid-10M and 1M Chinese ones from Youku-mPLUG-10M. By substituting WebVid-10M with 1M refined video-text pairs, we reduce 90$\%$ of data volume while achieving a noticeable performance gain of +1.7 at \textit{MSRVTT-AVG-R}.

% -----------------------------------

\subsection{Step 2: Using Key-frames as Video Inputs}
\label{method:step-2}

Key-frames aim to maximize the diversity among extracted frames, which highly represent the video content while skipping the time-consuming video decoding step. We conduct a contrast experiment that only replacing raw video inputs with key-frames under the same setting. Results demonstrate that such a simple replacement yields a 56$\%$ of total time reduction and a performance gain of +1.5 at \textit{MSRVTT-AVG-R}. It verifies that key-frames are more effective and cost-efficient video inputs for pre-training.

% -----------------------------------

\subsection{Step 3: Adding Temporal Modeling Modules}
\label{method:step-3}

After addressing challenges of data quality and video input types, we turn to the issues of model design. As aforementioned, temporal modeling is essential for video-text pre-training, as some video descriptions necessitate a holistic understanding by exploiting temporal cues within adjacent frames. In this paper, we employ STAN \cite{liu2023revisiting} as an example for advanced temporal modeling.

As illustrated in the middle-center part of Figure \ref{fig:framework}, STAN adopts a branch structure with decomposed spatial-temporal modules to enable temporal modeling. Specifically, STAN consists of a stack of $K$ spatial-temporal layers. Regarding the forward process of each layer, STAN first feeds the input features into a temporal self-attention module. STAN then exploits the multi-head self-attention mechanism within the visual encoder layer to construct its spatial self-attention module. Ultimately, STAN combines the outputs of the last visual encoder layer and the last STAN layer, generating video features that contain rich temporal information.

Due to limited space, we omit the detailed derivation and equations of STAN, while elaborating on them in the supplementary material (Sec. \ref{sec:supple-stan}). Note that STAN is not our technical contributions. Readers can refer to \cite{liu2023revisiting} and \cite{liu2023mug} for more details. By integrating STAN to strengthen temporal modeling, we observe a performance gain of +1.1 at \textit{MSRVTT-AVG-R} compared with the baseline in Step 2.

% -----------------------------------

\subsection{Step 4: Enhancing Video Features}
\label{method:step-4}

We next study the impact of video feature enhancement strategies. As aforementioned, the partial misalignment phenomenon occurs quite commonly in many video-text pairs, which would hinder the pre-training process as those inconsistent frames bring non-negligible noise. With the intuition that not all frames contribute equally to video description understanding, video feature enhancement strategies aim to emphasize those well-paired frame features while suppressing the rest inconsistent ones. In this paper, we employ a mutual-guided (Mug) \cite{liu2023mug} alignment head as an implicit enhancement strategy, while proposing a novel auxiliary-caption-guided (ACG) module as an explicit one.

\subsubsection{Mutual-guided Alignment Head (Mug)}
\label{method:step-mug}

As illustrated in the top-right part of Figure \ref{fig:framework}, Mug attempts to filter out misaligned information among cross-modal features in an implicit manner. 

Specifically, Mug first derives a dot-product similarity matrix according to video features $\mathbf{V} \in {\mathbb{R}}^{{N}_{f}*d}$ and text features $\mathbf{T} \in {\mathbb{R}}^{{N}_{t}*d}$. Mug then assigns a frame-to-token attention score ${z}_{i,j}$ to each text token features $\mathbf{t}_{j}$ based on its relevance to the current video frame features $\mathbf{v}_{i}$. Afterward, for each frame, Mug aggregates initial text features based on the attention distribution, yielding frame-specific text features $\hat{\mathbf{t}}_{i} = \sum_{j=1}^{{N}_{t}}({z}_{i,j}\mathbf{t}_{j})$. Mug then measures the consistency of each frame features $\mathbf{v}_{i}$ with respect to the text tokens $\hat{\mathbf{t}}_{i} \in \mathbb{R}^{d}$. In this way, Mug aggregates all frame-wise features, obtaining the enhanced text-guided video features $\Bar{\Bar{\mathbf{v}}} = \sum_{i=1}^{{N}_{f}}(\hat{z}_{i}\mathbf{v}_{i})$. Similarly, the process of generating enhanced video-guided text features $\Bar{\Bar{\mathbf{t}}}$ is a mirror operation as $\Bar{\Bar{\mathbf{v}}}$. 

Due to limited space, we omit the detailed derivation and equations of Mug, while elaborating on them in the supplementary material (Sec. \ref{sec:supple-mug}). Note that Mug is not our technical contributions. Readers can refer to \cite{liu2023mug} for more details.

\subsubsection{Auxiliary-Caption-Guided Module (ACG)}
\label{method:ACG}

As illustrated in the top-center part of Figure \ref{fig:framework}, ACG is devised to enhance video features in an explicit manner. 

Specifically, ACG exploits auxiliary frame captions $a$ provided by intermediate results of the refined dataset in Step 1 (Sec. \ref{method:step-1}). We first feed frame captions $a$ into the text encoder ${E}_{txt}$, obtaining [\textit{CLS}] token features as auxiliary caption features $\textbf{C} \in \mathbb{R}^{{N}_{f} * d}$.

We then assign an explicit weight to each frame features ${\mathbf{v}}_{i}$ following two strategies: Frame-Caption Re-weighting (FCR) and Text-Caption Re-weighting (TCR). FCR and TCR encourage the contribution of well-aligned frame features and suppress inconsistent ones from inter-modal and intra-modal aspects:

1) FCR utilizes the inter-modal consistency score ${s}_{i}^{v}$ to measure whether a given frame-caption features pair $({\mathbf{v}}_{i}, {\mathbf{c}}_{i})$ is well-paired, which could be formulated as:
\begin{equation}
\label{eq:acg-fcr}
	{s}_{i}^{v} = \frac{ {\rm exp}^{\lambda \langle \mathbf{c}_{i} , \mathbf{v}_{i} \rangle}}{ \sum_{i=1}^{{N}_{f}} {\rm exp}^{\lambda \langle \mathbf{c}_{i}, \mathbf{v}_{i} \rangle}},
\end{equation}
where $\lambda$ controls the sharpness of the score distribution in ACG.

2) TCR dynamically adjusts the weights of each frame depending on the intra-modal similarity score ${s}_{i}^{t}$, which is computed between auxiliary caption features ${\mathbf{c}}_{i}$ and target text features ${\mathbf{t}}_{cls}$ as:
\begin{equation}
\label{eq:acg-fcr}
	{s}_{i}^{t} = \frac{ {\rm exp}^{\lambda \langle \mathbf{c}_{i} , \mathbf{t}_{cls} \rangle}}{ \sum_{i=1}^{{N}_{f}} {\rm exp}^{\lambda \langle \mathbf{c}_{i} , \mathbf{t}_{cls} \rangle}}.
\end{equation}

In this way, we employ inter-modal scores ${s}_{i}^{v}$ and intra-modal scores ${s}_{i}^{t}$ to obtain enhanced video features $\tilde{\mathbf{v}}_{i}$ as:
\begin{equation}
\label{eq:acg-fcr}
    \tilde{\mathbf{v}}_{i} = \frac{({s}_{i}^{v}+{s}_{i}^{t})}{2}\mathbf{v}_{i}.
\end{equation}

Besides, we also calculate the Frame-Caption Contrastive loss $\mathcal{L}_{fcc}$ to preserve well-learned knowledge embedded in robust image-text models, which could be formulated as:
\begin{equation}
\label{eq:FCC}
	\mathcal{L}_{fcc}(\mathbf{v}, \mathbf{f}) = -\frac{1}{{N}_{f}} \sum_{i=1}^{{N}_{f}} \sum_{b=1}^{\mathcal{B}}  {\rm log}   \frac{{\rm exp}^{ \langle {\mathbf{v}}_{i}^{b}, \mathbf{c}_{i}^{b} \rangle} }
	{ {\rm exp}^{ \langle {\mathbf{v}}_{i}^{b}, {\mathbf{c}}_{i}^{b} \rangle} + \sum_{e \neq b}{\rm exp}^{ \langle {\mathbf{v}}_{i}^{e}, {\mathbf{c}}_{i}^{b} \rangle} }.
\end{equation}

Note that the output video features $\tilde{\mathbf{V}} = \{\tilde{\mathbf{v}}_{i}\}_{i=1}^{{N}_{f}}$ of ACG would feed as the inputs of the Mug head, \textit{i}.\textit{e}., $\Bar{\Bar{\mathbf{v}}}, \Bar{\Bar{\mathbf{t}}} = {\rm Mug} (\tilde{\mathbf{V}}, \mathbf{T})$. The overall object function in our final model could be formulated as:
\begin{equation}
\label{eq:VTC}
    \mathcal{L}_{all} = \mathcal{L}_{vtc}(\Bar{\Bar{\mathbf{v}}}, \Bar{\Bar{\mathbf{t}}}) + \mathcal{L}_{fcc}(\mathbf{v}, \mathbf{f}).
\end{equation}

By leveraging implicit strategy Mug and explicit strategy ACG to enhance video features, we obtain an obvious performance gain of +3.2 at \textit{MSRVTT-AVG-R}.

% -----------------------------------

\subsection{Other Useful Designs to Reach the SOTA}
\label{method:trick}

Here, we enumerate three extra designs that help our M$^{2}$-RAAP reach a new SOTA. We only employ these designs in Table \ref{result:all}. We leave it as a future work to further explore their inner contributions.

\textbf{Dual Softmax Loss (DSL)}. We employ DSL \cite{cheng2021improving} only in the evaluation step to enhance video and text features, whose calculation is: $\Bar{\Bar{\mathbf{v}}} = \Bar{\Bar{\mathbf{v}}} \cdot {\rm softmax}(\Bar{\Bar{\mathbf{v}}})$, $\Bar{\Bar{\mathbf{t}}} = \Bar{\Bar{\mathbf{t}}} \cdot {\rm softmax}(\Bar{\Bar{\mathbf{t}}})$. Readers could refer to \cite{cheng2021improving} for more details.

\textbf{Post-pre-training on image-text datasets}. We first post-pre-train CLIP on an image dataset, CC2M, whose parameters are utilized to initial our model at the start of video-text pre-training.

\textbf{Pre-training with mixed re-written captions}. We mix up long and short re-written text annotations at the rate of 1:1 in Step 1, enhancing the generalization capability of video-text models.

%%%%%%%%%%%%%%% Experiments %%%%%%%%%%%%%%%

\begin{table*}[t]
	\small
	\centering
	\begin{tabular}{p{2.4cm}<{\centering}|p{0.8cm}<{\centering}|p{3.0cm}<{\centering}|p{3.0cm}<{\centering}|p{3.0cm}<{\centering}|p{3.0cm}<{\centering}}
		\hline
		\multicolumn{1}{c|}{\multirow{2}{*}{Method}} &
		\multicolumn{1}{c|}{Data} &
		\multicolumn{1}{c|}{MSRVTT} &
		\multicolumn{1}{c|}{DiDemo} &
		\multicolumn{1}{c|}{LSMDC} &
  	\multicolumn{1}{c}{Activity-Net}
		\\ \cline{3-6} 
		&
		\multicolumn{1}{c|}{Pairs} &
		\multicolumn{1}{c|}{R@1/5/10 (AVG-R)} &
		\multicolumn{1}{c|}{R@1/5/10 (AVG-R)} &
		\multicolumn{1}{c|}{R@1/5/10 (AVG-R)} &
		\multicolumn{1}{c}{R@1/5/10 (AVG-R)}
		\\ \hline
		% CLIP-L/14 & 10M & 35.9 / 60.8 / 69.6 (55.4) & 36.2 / 62.3 / 71.1 (56.5) & 18.0 / 33.3 / 41.4 (30.9) & - & -
		% \\ % \hline
  
        \multicolumn{6}{c}{\multirow{1}{*}{ \textit{Methods: Video-text Pre-training from Scratch}}} \\\hline
        Singularity & 17M & 34.0 / 56.7 / 66.7 (52.5) & 37.1 / 61.7 / 69.9 (56.2) & - & 30.6 / 55.6 / 66.9 (51.0)
		\\ % \hline
        HiTeA & 17M & 34.4 / 60.0 / 69.9 (54.8) & 43.2 / 69.3 / 79.0 (63.8) & 18.3 / 36.7 / 44.2 (33.1) & -
		\\ % \hline
        HiVLP & 116M & 43.5 / 66.4 / 76.4 (62.1) & - & - & -
		\\ \hline

        \multicolumn{6}{c}{\multirow{1}{*}{ \textit{Methods: Video-text Pre-training by Adapting Image-text Models}}} \\\hline
        Intern-Vid & 12M & 40.0 / 65.3 / 74.1 (59.8) & 31.5 / 57.6 / 68.2 (52.4) & 17.6 / 32.4 / 40.2 (30.1) & -
		\\ % \hline
        BT-Adapter & 2.5M & 40.9 / 64.7 / 73.5 (59.7) & 35.6 / 61.9 / 72.6 (56.7) & 19.5 / 35.9 / 45.0 (33.5) & 37.0 / 66.7 / 78.9 (60.9)
		\\ % \hline
        Mug-STAN & 10M & 41.7 / 65.7 / 75.8 (61.1) & 39.6 / 64.3 / 72.6 (58.8) & 20.7 / 38.8 / 46.2 (35.2) & -
        \\
        UMT-L & 25M & 40.7 / 63.4 / 71.8 (58.6) & \textbf{48.6} / \textbf{72.9} / 79.0 (\textbf{66.8}) & \textbf{24.9} / 41.7 / 51.8 (39.5) & \textbf{41.9} / 68.9 / 80.3 (63.7)
		\\ \hline

        \multicolumn{6}{c}{\multirow{1}{*}{ \textit{Methods: Video-text Pre-training is part of Large Multi-modal Models}}} \\\hline
        VideoCoCa & 108M & 34.3 / 57.8 / 67.0 (53.0) & - & - & 34.5 / 63.2 / 76.6 (58.1)
        \\
        Lang-Bind & 10M & 42.6 / 65.4 / 75.5 (61.2) & 37.8 / 63.2 / 73.4 (58.1) & - & 35.1 / 63.4 / 76.6 (58.4)
		\\ % \hline
        mPLUG-2 & 17M & \textbf{47.1} / 69.7 / 79.0 (65.3) & 45.7 / 71.1 / \textbf{79.2} (65.3) & 24.1 / 43.8 / 52.0 (40.0) & -
		% \hline
		\\ \hline
		% \multicolumn{7}{c}{\multirow{+1.5}{*}{ \textit{M$^{2}$-RAAP}}}
		% \vspace{0.5em} \\\hline

        \multicolumn{6}{c}{\multirow{1}{*}{ \textit{Our M$^{2}$-RAAP Recipe}}} \\\hline
        \rowcolor{mygray} M$^{2}$-RAAP-CLIP$^{*}$ & 1M & 46.1 / \textbf{70.4} / \textbf{80.6} (\textbf{65.7}) & 43.6 / 68.9 / 77.3 (63.2) & 24.4 / \textbf{44.3} / \textbf{52.0} (\textbf{40.2}) & 41.2 / \textbf{70.2} / \textbf{82.4} (\textbf{64.6})
        \\\hline

	\end{tabular}

\caption{Performance comparison of different methods on four English downstream zero-shot video-text retrieval datasets. M$^{2}$-RAAP establishes a new SOTA on four of them. The superscript ``*'' denotes that M$^{2}$-RAAP employs extra designs in Sec. \ref{method:trick}.}
\label{result:all}
\end{table*}

\begin{table}[t]
	\small
	\centering
	\begin{tabular}{p{2.4cm}<{\centering}|p{0.8cm}<{\centering}|p{3.0cm}<{\centering}}
		\hline
		\multicolumn{1}{c|}{\multirow{2}{*}{Method}} &
		\multicolumn{1}{c|}{Data} &
		\multicolumn{1}{c}{VATEX-English} 
		\\ \cline{3-3} 
		&
		\multicolumn{1}{c|}{Pairs} &
		\multicolumn{1}{c}{R@1/5/10 (AVG-R)}
		\\ \hline
		
        Intern-Vid & 12M & 49.5 / - / - ( - )
	\\\hline
        VideoCoCa & 108M & 53.2 / 83.3 / 90.1 (75.5)
        \\\hline
        \rowcolor{mygray} M$^{2}$-RAAP-CLIP$^{*}$ & 1M & \textbf{58.0} / \textbf{86.3} / \textbf{92.6} (\textbf{79.0})
        \\\hline

	\end{tabular}
\caption{Performance comparison on VATEX-English.}
\label{result:all-vatex}
\end{table}

\section{Experiments}
\label{experiments}

\subsection{Datasets and Basic Image-text Models}

\subsubsection{Video-text Pre-training Datasets} 

To validate the reliability and portability of our M$^{2}$-RAAP recipe, we utilize two widely-employed datasets: English WebVid-10M \cite{bain2021frozen} and Chinese Youku-mPLUG-10M \cite{xu2023youku}. Following our data filtering and refinement pipeline in Sec. \ref{method:step-1}, our pre-training corpus includes 1M English high-quality refined video-text pairs (dubbed WebVid-Refined-1M) and 1M Chinese ones (dubbed Youku-Refined-1M).

\subsubsection{Zero-shot Video-text Retrieval Datasets} 

Zero-shot video-text retrieval tasks require pre-trained models to directly inference on the testing set without additional fine-tuning, which fairly measures the generalization capability of video-text models. In this paper, we introduce five widely-employed English video-text retrieval testing sets and two Chinese ones, to thoroughly compare our performance with various current baselines. 

For English zero-shot retrieval testing sets, they are: 

% 1) \textit{MSRVTT} \cite{xu2016msr} with 1K video-text pairs, 2) \textit{DiDemo} \cite{anne2017localizing} with 1K video-text pairs, 3) \textit{LSMDC} \cite{rohrbach2017movie} with 1K video-text pairs, 4) \textit{Activity-Net} \cite{krishna2017dense} with 4.9K video-text pairs, and 5) \textit{VATEX-English} \cite{wang2019vatex} is a multi-choice retrieval dataset with 1.4K videos and 14K associated text.

\begin{itemize}
\setlength{\itemsep}{0pt}
\setlength{\parsep}{0pt}
\setlength{\parskip}{0pt}

	\item \textit{MSRVTT} \cite{xu2016msr} with 1K video-text pairs;
	
	\item \textit{DiDemo} \cite{anne2017localizing} with 1K video-text pairs;
	
	\item \textit{LSMDC} \cite{rohrbach2017movie} with 1K video-text pairs;
	
	\item \textit{Activity-Net} \cite{krishna2017dense} with 4.9K video-text pairs;

        \item \textit{VATEX-English} \cite{wang2019vatex} is a multi-choice retrieval dataset with 1.4K videos and 14K associated text.
\end{itemize}

For Chinese zero-shot retrieval testing sets, they are: 

% 1) \textit{VATEX-Chinese} \cite{wang2019vatex} with 1.4K video-text pairs, and 2) \textit{Youku-Retrieval} \cite{xu2023youku} with 1.7K video-text pairs.

\begin{itemize}
\setlength{\itemsep}{0pt}
\setlength{\parsep}{0pt}
\setlength{\parskip}{0pt}

	\item \textit{VATEX-Chinese} \cite{wang2019vatex} with 1.4K video-text pairs;
	
	\item \textit{Youku-Retrieval} \cite{xu2023youku} with 1.7K video-text pairs.
	
\end{itemize}

\noindent\textbf{Metrics.} We employ Recall@K (R@K, K=1/5/10) and their average results (AVG-R) to measure zero-shot retrieval performance, where AVG-R = (R@1+R@5+R@10)/3.0. We denote the metric of AVG-R on the MSRVTT dataset as \textit{MSRVTT-AVG-R}, which is used to quantitatively assess the performance gain in the following section.

\subsubsection{Basic Image-text Models}

We implement our M$^{2}$-RAAP recipe on three basic image-text models to ensure the reliability of our conclusions, \textit{i}.\textit{e}., 1) CLIP (we use the CLIP-ViT-L/14 version) \cite{radford2021learning}, AltCLIP \cite{chen2022altclip}, and M$^{2}$-Encoder \cite{guo2024m2encoder} (we use the 1B version). Specifically, we adapt CLIP and AltCLIP on English WebVid-Refined-1M, and adapt M$^{2}$-Encoder on Chinese Youku-Refined-1M.

% -----------------------------------

\begin{table*}[t]
    \small
	\centering
	\begin{tabular}{p{0.4cm}<{\centering}|p{1.2cm}<{\centering}|p{1.2cm}<{\centering}|p{1.2cm}<{\centering}|p{1.2cm}<{\centering}|p{1.6cm}<{\centering}|p{3.0cm}<{\centering}
	|p{3.0cm}<{\centering}}
		\hline
		\multicolumn{1}{c|}{\multirow{2}{*}{No.}} &
		\multicolumn{1}{c|}{Basic} &
		\multicolumn{1}{c|}{WebVid} &
        \multicolumn{1}{c|}{Video} &
		\multicolumn{1}{c|}{Temporal} &
		\multicolumn{1}{c|}{Enhancing} &
		\multicolumn{1}{c|}{MSRVTT} &
		\multicolumn{1}{c}{DiDemo}
		\\ \cline{7-8} 
		&
		\multicolumn{1}{c|}{Model} &
		\multicolumn{1}{c|}{Volumn} &
        \multicolumn{1}{c|}{Inputs} &
		\multicolumn{1}{c|}{Module} &
		\multicolumn{1}{c|}{Strategy} &
		\multicolumn{1}{c|}{R@1/5/10 (AVG-R)} &
		\multicolumn{1}{c}{R@1/5/10 (AVG-R)}
		\\ \hline
  
		\multicolumn{8}{c}{\multirow{1}{*}{ \textit{Adapting Model: CLIP}}}
		\\ \hline
  
		A0 & CLIP & 10M & raw-vid & - & - & 38.9 / 63.4 / 72.5 (58.3) & 35.8 / 59.4 / 69.4 (54.9)
		\\ % \hline
		A1 & CLIP & ref-1M & raw-vid & - & - & 41.6 / 64.5 / 73.8 (60.0) & 36.3 / 60.9 / 71.7 (56.3)
		\\ % \hline
		A2 & CLIP & ref-1M & key-frm & - & - & 42.1 / 66.1 / 76.3 (61.5) & 39.2 / 63.8 / 73.7 (58.9)
		\\ % \hline
		A3 & CLIP & ref-1M & key-frm & STAN & - & 43.6 / 67.7 / 76.4 (62.6) & 40.6 / 64.0 / 73.1 (59.2)
        \\ % \hline
        \rowcolor{mygray} A4 & CLIP & ref-1M & key-frm & STAN & Mug + ACG & \textbf{46.0} / \textbf{71.4} / \textbf{80.0} (\textbf{65.8}) & \textbf{42.8} / \textbf{66.9} / \textbf{75.7} (\textbf{61.8})
    	\\ \hline
     
		\multicolumn{8}{c}{\multirow{1}{*}{ \textit{Adapting Model: AltCLIP}}}
		\\\hline
  
		B0 & AltCLIP & 10M & raw-vid & - & - & 40.3 / 65.7 / 74.6 (60.2) & 30.8 / 54.6 / 65.0 (50.1)
		\\ % \hline
		B1 & AltCLIP & ref-1M & raw-vid & - & - & 40.8 / 65.7 / 75.3 (60.6) & 33.6 / 56.0 / 65.5 (51.7)
		\\ % \hline
		B2 & AltCLIP & ref-1M & key-frm & - & - & 43.8 / 66.4 / 76.5 (62.2) & 32.1 / 56.3 / 64.8 (51.1)
		\\ % \hline
		B3 & AltCLIP & ref-1M & key-frm & STAN & - & 44.4 / 67.7 / 76.8 (63.0) & 34.5 / 56.7 / 64.5 (51.9)
        \\ % \hline
        \rowcolor{mygray} B4 & AltCLIP & ref-1M & key-frm & STAN & Mug + ACG & \textbf{47.3} / \textbf{70.9} / \textbf{79.4} (\textbf{65.9}) & \textbf{35.5} / \textbf{58.9} / \textbf{67.9} (\textbf{54.1})
    	\\ \hline
		
	\end{tabular}
\caption{Ablation study of our M$^{2}$-RAAP recipe. M$^{2}$-RAAP starts with a robust image-text model, and employs a progressive expansion scheme to evaluate the rewards of four components. All experiments are under the same setting without designs in Sec. \ref{method:trick}.
% improving data quality, video input types, temporal modeling, and video feature enhancement.
}
\label{ablation_M$^{2}$-RAAP}
\end{table*}

\begin{table*}[t]
    \small
    \centering
	\begin{tabular}{p{0.4cm}<{\centering}|p{1.8cm}<{\centering}|p{1.2cm}<{\centering}|p{1.0cm}<{\centering}|p{0.6cm}<{\centering}p{0.6cm}<{\centering}p{0.6cm}<{\centering}|p{3.0cm}<{\centering}|p{3.0cm}<{\centering}}
		\hline
		\multicolumn{1}{c|}{\multirow{2}{*}{No.}} &
		\multicolumn{1}{c|}{Basic} &
		\multicolumn{1}{c|}{WebVid} &
        \multicolumn{1}{c|}{Using} &
		\multicolumn{3}{c|}{ACG Strategy} &
		\multicolumn{1}{c|}{MSRVTT} &
		\multicolumn{1}{c|}{DiDemo}
		\\ \cline{5-9} 
		&
		\multicolumn{1}{c|}{Model} &
		\multicolumn{1}{c|}{Volumn} &
        \multicolumn{1}{c|}{Key-frm} &
		\multicolumn{1}{c}{FCC} &
        \multicolumn{1}{c}{FCR} &
        \multicolumn{1}{c|}{TCR} &
		\multicolumn{1}{c|}{R@1/5/10 (AVG-R)} &
		\multicolumn{1}{c|}{R@1/5/10 (AVG-R)}
  
		\\ \hline
		\multicolumn{9}{c}{\multirow{1}{*}{ \textit{Adapting Model: CLIP with Mug-STAN}}}
  
		\\\hline
		C1 & CLIP-M-S & 10M & - & - & - & - & 42.1 / 65.4 / 74.1 (60.5) & 38.7 / 64.3 / 72.8 (58.6)
		\\ \hline
		C2 & CLIP-M-S & ref-1M & - & - & - & - & 44.6 / 67.7 / 76.9 (63.1) & 39.5 / 63.7 / 73.0 (58.7)
		\\ \hline
		C3 & CLIP-M-S & ref-1M & \footnotesize{$\surd$} & - & - & - & 44.7 / 70.7 / 77.3 (64.2) & 42.0 / 66.7 / 75.2 (61.3)
        \\ \hline
        C4 & CLIP-M-S & ref-1M & \footnotesize{$\surd$} & \footnotesize{$\surd$} & - & - & 45.5 / 71.1 / 79.1 (65.2) & 42.3 / \textbf{67.4} / 75.3 (61.6)
        \\ % \hline
        C5 & CLIP-M-S & ref-1M & \footnotesize{$\surd$} & \footnotesize{$\surd$} & \footnotesize{$\surd$} & - & 45.0 / \textbf{72.0} / 79.2 (65.4) & 41.2 / 66.8 / 74.2 (60.7)
    	\\ % \hline
        C6 & CLIP-M-S & ref-1M & \footnotesize{$\surd$} & \footnotesize{$\surd$} & - & \footnotesize{$\surd$} & \textbf{46.4} / 70.9 / 79.3 (65.5) & 42.2 / 66.3 / 73.3 (60.6)
    	\\ \hline
     
        \rowcolor{mygray} C7 & CLIP-M-S & ref-1M & \footnotesize{$\surd$} & \footnotesize{$\surd$} & \footnotesize{$\surd$} & \footnotesize{$\surd$} & 46.0 / 71.4 / \textbf{80.0} (\textbf{65.8}) & \textbf{42.8} / 66.9 / \textbf{75.7} (\textbf{61.8})
    	\\ \hline
     
		\multicolumn{9}{c}{\multirow{1}{*}{ \textit{Adapting Model: AltCLIP with Mug-STAN}}}\\\hline
  
		D1 & AltCLIP-M-S & 10M & - & - & - & - & 43.6 / 67.4 / 76.3 (62.4) & 33.3 / 56.1 / 65.1 (51.5)
		\\ \hline
		D2 & AltCLIP-M-S & ref-1M & - & - & - & - & 44.2 / 68.5 / 76.6 (63.1) & 35.1 / 58.2 / 67.2 (53.5)
		\\ \hline
		D3 & AltCLIP-M-S & ref-1M & \footnotesize{$\surd$} & - & - & - & 46.2 / 70.2 / 78.2 (64.9) & 33.5 / \textbf{59.2} / 67.1 (53.3)
		\\ \hline
		D4 & AltCLIP-M-S & ref-1M & \footnotesize{$\surd$} & \footnotesize{$\surd$} & - & - & 46.3 / 69.9 / 79.4 (65.2) & 34.3 / 58.9 / 66.7 (53.3)
        \\ % \hline
        D5 & AltCLIP-M-S & ref-1M & \footnotesize{$\surd$} & \footnotesize{$\surd$} & \footnotesize{$\surd$} & - & 46.8 / 70.2 / \textbf{79.6} (65.5) & 35.0 / 58.2 / 67.6 (53.6)
        \\ % \hline
        D6 & AltCLIP-M-S & ref-1M & \footnotesize{$\surd$} & \footnotesize{$\surd$} & - & \footnotesize{$\surd$} & 46.7 / 70.8 / 79.2 (65.6) & \textbf{35.7} / 58.2 / 67.2 (53.7)
    	\\ \hline
        \rowcolor{mygray} D7 & AltCLIP-M-S & ref-1M & \footnotesize{$\surd$} & \footnotesize{$\surd$} & \footnotesize{$\surd$} & \footnotesize{$\surd$} & \textbf{47.3} / \textbf{70.9} / 79.4 (\textbf{65.9}) & 35.5 / 58.9 / \textbf{67.9} (\textbf{54.1})
    	\\ \hline
		
	\end{tabular}
 
  \caption{Ablation study of our technical contributions, including refining text annotations, replacing video inputs with key-frames, and introducing ACG to enhance video features. All experiments employ Mug-STAN (M-S) without designs in Sec. \ref{method:trick}.}
\vspace{-0.0cm}
\label{ablation_novelty}
\end{table*}

\subsection{Experimental Settings}
\label{experimental-setting}

\subsubsection{Uniform Experimental Settings}

Regarding pre-training implementation details, we process raw videos into 8 (${N}_{f}$) frames by uniform sampling or key-frame extracting. We set the length of text tokens ${N}_{t} = 64$, and the input frame size is 224 $*$ 224. All models are pre-trained by the AdamW optimizer with a weight decay rate of 0.05. We set the initial learning rate to 2e-6 and conduct a cosine annealing decay schedule. All adaptation-based video-text pre-training lasts for 2 epochs. The batch size is 120 for CLIP, 96 for AltCLIP, and 120 for M$^{2}$-Encoder. The whole pre-training time of our final model is 92 GPU hours on NVIDIA A100. In total, it consumes 11.5 hours to pre-train on 1M data with 8 A100 GPUs.

Regarding the evaluation on zero-shot retrieval tasks, following the conventional protocol, we sample 12 frames for each video and set the length of text tokens ${N}_{t} = 77$.

\subsubsection{Special Settings in M$^{2}$-RAAP Recipe}

Regarding the data filtering and refinement pipeline in Step 1 (Sec. \ref{method:step-1}), we employ CLIP to compute similarity scores for English video-text pairs while utilizing M$^{2}$-Encoder for Chinese video-text ones. We exploit temporal segment density peaks clustering (TSDPC) \cite{tang2023deep} algorithm to obtain video key-frames. We employ BLIP2 \cite{li2023blip} to derive key-frame captions, and utilize ChatGLM3-6B \cite{du2022glm, zeng2022glm} to generate both long and short re-written video descriptions.

Regarding the temporal module STAN \cite{liu2023revisiting} in Step 3 (Sec. \ref{method:step-3}), we employ four STAN layers following the conventional protocol.

Regarding the video feature enhancement strategies in Step 4 (Sec. \ref{method:step-4}), we set the temperature scalar $\lambda=10$ for the proposed ACG strategy. The logit scale during pre-training is 100 by default.

Note that we employ the short re-written video descriptions to pre-train our models without extra designs in Sec. \ref{method:trick} by default.

% -----------------------------------

\subsection{Performance Comparison}

We compare the final model of our M$^{2}$-RAAP Recipe with various current state-of-the-art baselines, including the following three types of methods: 

1) Methods that pre-training from scratch: Singularity \cite{lei2022singularity}, HiTeA \cite{ye2023hitea}, and HiVLP \cite{shao2023hivlp}.

2) Methods that adapting image-text models: Intern-Vid \cite{wang2023internvid}, BT-Adapter \cite{liu2023bt-adapter}, Mug-STAN \cite{liu2023mug}, and UMT-L \cite{li2023unmasked}.

3) Methods that video-text pre-training is part of large multi-modal models: VideoCoCa \cite{yan2022videococa}, Lang-Bind \cite{zhu2023languagebind}, and mPLUG-2 \cite{xu2023mplug-2}.

Tables \ref{result:all} and \ref{result:all-vatex} presents detailed experimental results on five English downstream zero-shot video-text retrieval datasets, where we attach extra designs provided by Sec. \ref{method:trick} to our final model (M$^{2}$-RAAP-CLIP$^{*}$). Several conclusions could be reached as follows:

1) M$^{2}$-RAAP reaches a new SOTA at the metric of AVG-R on four datasets, outperforming current SOTA baselines by +0.4 on MSRVTT (mPLUG-2), +0.2 on LSMDC (mPLUG-2), +0.9 on Activity-Net (UMT-L), and +3.5 on VATEX-English (VideoCoCa). 

2) M$^{2}$-RAAP achieves a superior balance between performance and efficiency. M$^{2}$-RAAP only requires 11.5 hours for pre-training on 1M video-text pairs with 8 A100 GPUs (92h), which is much more efficient than current SOTA baselines. In contrast, UMT-L needs 130 hours to pre-train on 25M data pairs with 32 A100 GPUs (4160h).

3) M$^{2}$-RAAP even exhibits better performance compared with some large multi-modal models like mPLUG-2 and Language-Bind, indicating the effectiveness of our proposed recipe.

% -----------------------------------

\subsection{Ablation Study of M$^{2}$-RAAP Recipe}

\begin{figure}[t]
	\begin{center}
		\includegraphics[width=0.47\textwidth]{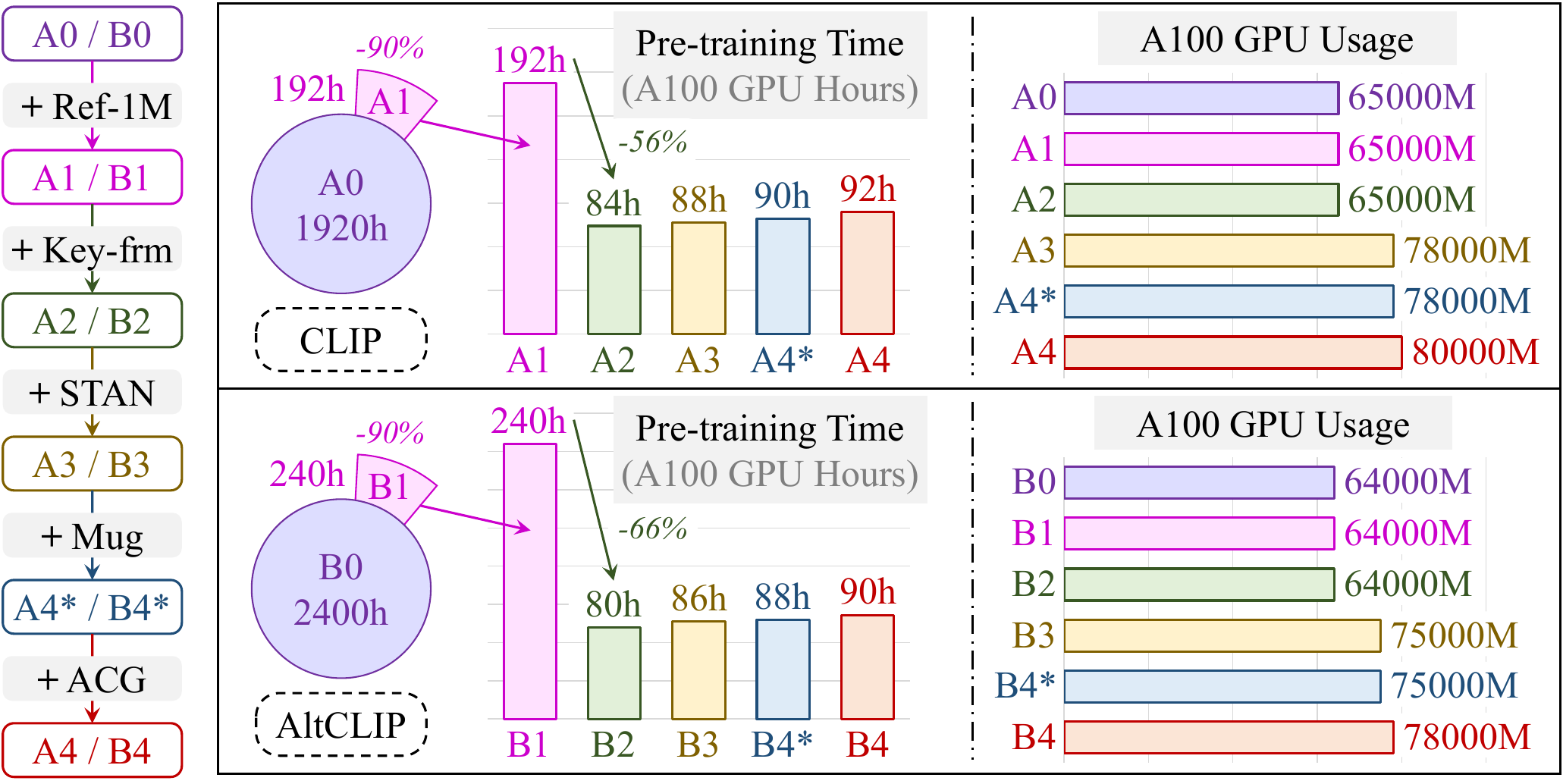}
	\end{center}
	\vspace{-0.4cm}
	\caption{Quantitative efficiency analysis towards the total pre-training GPU hours and GPU usage of M$^{2}$-RAAP.}
	
	\label{fig:effe}
\end{figure}

As aforementioned, M$^{2}$-RAAP adopts a progressive expansion scheme started with a robust image-text model. To ensure the reliability and portability of our M$^{2}$-RAAP recipe, we replicate M$^{2}$-RAAP on two image-text models, \textit{i}.\textit{e}., CLIP and AltCLIP. We ablate M$^{2}$-RAAP by gradually implementing the operation from Step 1 to Step 4 in Sec. \ref{method:all}, whose performance and efficiency rewards are illustrated in Table \ref{ablation_M$^{2}$-RAAP} and Figure \ref{fig:effe}. Several conclusions could be reached as follows:

1) Refining the noisy data corpus yields remarkable performance and efficiency gain. By substituting conventional WebVid-10M with our newly refined WebVid-Refined-1M, we reduce 90$\%$ of pre-training data (10M $\rightarrow$ 1M) while achieving a performance gain of +1.7 on CLIP (\textit{A0} $\rightarrow$ \textit{A1}) and +0.4 on AltCLIP (\textit{B0} $\rightarrow$ \textit{B1}) at \textit{MSRVTT-AVG-R}. It indicates the effectiveness of our developed data filtering and text re-writing pipeline.

2) Key-frames are more effective and cost-efficient inputs compared with raw videos. We observe that models employing key-frames outperform baselines by +1.5 on CLIP (\textit{A1} $\rightarrow$ \textit{A2}) and +1.6 on AltCLIP (\textit{B1} $\rightarrow$ \textit{B2}) at \textit{MSRVTT-AVG-R}, while halving the whole pre-training time on CLIP (-56$\%$). It encourages future research to prioritize the use of key-frames over raw videos during pre-training.
% , since the latter currently dominates as the default input for video-text pre-training.

3) Temporal modeling mechanisms are crucial for models to grasp the theme of a video from a holistic perspective. By employing STAN to achieve advanced temporal modeling, performance results at \textit{MSRVTT-AVG-R} increase by +1.1 on CLIP (\textit{A2} $\rightarrow$ \textit{A3}) and +0.8 on AltCLIP (\textit{B2} $\rightarrow$ \textit{B3}). 
% Note that current video-text datasets are mainly short videos, enhancing temporal modeling would be promising for future long video pre-training.

4) Enhancing video features remarkably alleviates the partially misaligned problem. By exploiting the implicit video feature enhancement strategy Mug and the explicit one ACG, we outperform baselines by +3.2 on CLIP (\textit{A3} $\rightarrow$ \textit{A4}) and +2.9 on AltCLIP (\textit{B3} $\rightarrow$ \textit{B4}) at \textit{MSRVTT-AVG-R}. It motivates us to keep exploring this direction.

In summary, our final models \textit{A4} and \textit{B4} achieve a total performance gain of +7.5 on CLIP and +5.7 on AltCLIP at \textit{MSRVTT-AVG-R}, with only 10$\%$ of total video-text data and 5$\%$ of pre-training time compared with the baselines \textit{A0} and \textit{B0}. It indicates that M$^{2}$-RAAP is an effective and efficient multi-modal recipe for zero-shot video-text retrieval. 

\begin{table}[t]
    \small
	\centering
	\begin{tabular}{p{2.1cm}<{\centering}|p{2.7cm}<{\centering}|p{2.7cm}<{\centering}}
		\hline
		\multicolumn{1}{c|}{\multirow{2}{*}{Method}} &
        \multicolumn{1}{c|}{VATEX-Chinese} &
		\multicolumn{1}{c}{Youku-Retrieval}
		\\ \cline{2-3} 
		&
		\multicolumn{1}{c|}{R@1/5/10 (AVG-R)} &
		\multicolumn{1}{c}{R@1/5/10 (AVG-R)}
		\\ \hline
        AltCLIP & 36.8/59.8/66.7 (54.4) & 20.5/42.3/50.9 (37.9)
		\\ \hline
        CN-CLIP & 40.4/72.6/82.1  (65.0) & -
		\\ \hline
        M$^{2}$-Encoder & 44.3/76.8/84.3 (68.5) & 30.9/55.8/65.8 (50.8)
		\\ \hline
		Youku-mPLUG & - & 17.5/43.8/56.4 (39.2)
		\\ \hline
        \rowcolor{mygray} M$^{2}$-RAAP & \textbf{53.0}/\textbf{81.6}/\textbf{88.4} (\textbf{74.3}) & \textbf{37.5}/\textbf{65.8}/\textbf{74.9} (\textbf{59.4})
        \\ \hline
		
	\end{tabular}

\caption{Performance comparison of different methods on two downstream Chinese zero-shot video-text retrieval datasets, where M$^{2}$-RAAP establishes a new SOTA.}
% \vspace{-0.4cm}
\label{results_chinese}
\end{table}

\begin{table}[t]
    \small
    \vspace{0.0cm}
	\centering
	\begin{tabular}{p{0.4cm}<{\centering}|p{0.6cm}<{\centering}|p{0.6cm}<{\centering}|p{0.6cm}<{\centering}|p{3.0cm}<{\centering}}
		\hline
		\multicolumn{1}{c|}{\multirow{2}{*}{No.}} &
		\multicolumn{1}{c|}{Use} &
  	\multicolumn{1}{c|}{Key} &
        \multicolumn{1}{c|}{Use} &
        \multicolumn{1}{c}{VATEX-Chinese}
		\\ \cline{5-5} 
		&
		\multicolumn{1}{c|}{R1M} &
  	\multicolumn{1}{c|}{Frm} & 
        \multicolumn{1}{c|}{ACG} &
		\multicolumn{1}{c}{R@1/5/10 (AVG-R)}
		\\ \hline
		E1 & - & - & - & 49.8 / 78.9 / 86.5 (71.7)
		\\ \hline
		E2 & \footnotesize{$\surd$} & - & - & 51.0 / 81.3 / 88.4 (73.6)
		\\ \hline
		E3 & \footnotesize{$\surd$} & \footnotesize{$\surd$} & - & 51.9 / 81.2 / 88.1 (73.8)
        \\ \hline
        \rowcolor{mygray} E4 & \footnotesize{$\surd$} & \footnotesize{$\surd$} & \footnotesize{$\surd$} & \textbf{53.0} / \textbf{81.6} / \textbf{88.4} (\textbf{74.3})
        \\ \hline
		
	\end{tabular}

% \vspace{-0.4cm}
\caption{Ablation study of our technical contributions in Chinese pre-training. ``R1M'' denotes Youku-Refined-1M.}
\label{ablation_chinese}
\end{table}

% -----------------------------------

\subsection{Ablation Study of Technical Contributions}

As aforementioned, our technical contributions lie in the following three aspects: 1) the data filtering and text re-writing pipeline resulting in 1M high-quality bilingual video-text pairs (Sec. \ref{method:step-1}), 2) video inputs replacement with key-frames (Sec. \ref{method:step-2}), and 3) Auxiliary-Caption-Guided (ACG) video feature enhancement strategy (Sec. \ref{method:ACG}), where STAN and Mug are not our contributions. Therefore, we further ablate the above three contributions based on the Mug-STAN baseline with CLIP and AltCLIP. As illustrated in Table \ref{ablation_novelty}, several conclusions could be reached as follows:

1) The developed data filtering and text re-writing pipeline significantly enhances the data quality by excluding noisy video-text pairs and refining ambiguous annotations. In this way, we outperform baselines by +2.6 on CLIP (\textit{C1} $\rightarrow$ \textit{C2}) and +0.7 on AltCLIP (\textit{D1} $\rightarrow$ \textit{D2}) at \textit{MSRVTT-AVG-R}.

2) The replacement of video inputs with key-frames is a successful attempt, yielding a performance improvement of +1.1 on CLIP (\textit{C2} $\rightarrow$ \textit{C3}) and +1.8 on AltCLIP (\textit{D2} $\rightarrow$ \textit{D3}) at \textit{MSRVTT-AVG-R}.

3) ACG emerges as an effective video feature enhancement strategy to promote retrieval performance. We observe an obvious performance gain of +1.6 on CLIP (\textit{C3} $\rightarrow$ \textit{C7}) and +1.0 on AltCLIP (\textit{D3} $\rightarrow$ \textit{D7}) at \textit{MSRVTT-AVG-R}. Further ablation of the inner components in ACG indicates that: a) Frame-Caption Contrastive (FCC)  helps preserve well-learned knowledge embedded in image-text models, and 2) Frame-Caption Re-weighting (FCR) and Text-Caption Re-weighting (TCR) are complementary strategies to explicitly enhance video features from inter-modal and intra-model aspects. Besides, efficiency analysis in Figure \ref{fig:effe} indicates that ACG would not heavily increase the GPU usage or the time cost.

% -----------------------------------

\subsection{Results and Ablation Study in Chinese}

To further demonstrate the reliability of our M$^{2}$-RAAP recipe, we replicate three technical novelties on Chinese video-text pre-training and retrieval datasets. Performance comparison and ablation results are illustrated in Tables \ref{results_chinese} and \ref{ablation_chinese}, respectively. Specifically, Table \ref{results_chinese} shows that our models remarkably outperform current baselines. In Table \ref{ablation_chinese}, we find that our three technical contributions yield a notable performance gain on VATEX-Chinese.

Since the research on Chinese video-text pre-training is relatively nascent, it is difficult for us to choose representative and robust baselines. We hope that our work will be an opportunity to ignite more research interest in Chinese video-text pre-training.

%%%%%%%%%%%%%%%  Conclusion %%%%%%%%%%%%%%%

\section{Conclusion}

We present M$^{2}$-RAAP, an effective and efficient multi-modal recipe for improving zero-shot video-text retrieval by adapting pre-trained image-text models. M$^{2}$-RAAP adopts a progressive expansion scheme, targeting to qualitatively evaluate the performance and efficiency gain by improving data quality, replacing video inputs, adding temporal modeling, and enhancing video features. Following our M$^{2}$-RAAP recipe, we achieve a new SOTA on four English zero-shot testing sets and two Chinese ones, with only 10$\%$ of data volume and 5$\%$ of time cost. 
In the future, we aim to contain more candidate modules to refine our recipe while exploring new promising directions.
% while exploring new promising directions in adaptation-based video-text pre-training.

%%%%%%%%% REFERENCES
\newpage
{\small
\bibliographystyle{ieee_fullname}
\bibliography{egbib}
}
\newpage

\section{Supplementary Material}
\label{sec:supple}

In this supplementary material, we present detailed derivation and equations of STAN in \ref{method:step-3} and Mug in \ref{method:step-mug}.

\subsection{STAN in Step 3}
\label{sec:supple-stan}

As illustrated in the middle-center part of the overall framework in Figure \ref{fig:framework}, STAN adopts a branch structure with decomposed spatial-temporal modules to enable generalizable temporal modeling. Specifically, STAN consists of a stack of $K$ spatial-temporal layers. We first denote the outputs of the ${m}^{th}$ visual encoder layer as:
% \begin{equation}
% \label{eq:stan-1st}
% 	\mathcal{V}^{m} = \{\mathbf{P}_{i}^{m}\}_{i=1}^{{N}_{f}}, \quad \mathbf{P}_{i}^{m} =\{\mathbf{p}_{i,0}^{m}, \mathbf{p}_{i,1}^{m}, \cdots, \mathbf{p}_{i,{N}_{p}}^{m}\},
% \end{equation}
\begin{equation}
\label{eq:stan-1st}
	\mathcal{V}^{m} = \{\mathbf{P}_{i}^{m}\}_{i=1}^{{N}_{f}}, \quad \mathbf{P}_{i}^{m} =\{\mathbf{p}_{i,0}^{m}, \mathbf{p}_{i,1}^{m}, \cdots, \mathbf{p}_{i,{N}_{p}}^{m}\},
\end{equation}
where $\mathbf{p}_{i,0}^{m}$ represents the [\textit{CLS}] token of each frame features, ${N}_{p}$ denotes the number of per-frame patches. While for the inputs $\dot{\mathcal{V}}^{k}$ and outputs $\ddot{\mathcal{V}}^{k}$ of each STAN layer, we denote them as:
\begin{equation}
\label{eq:stan-1st}
	\dot{\mathcal{V}}^{k} = [\dot{\mathbf{p}}_{0,0}^{k}, \{\dot{\mathbf{P}}_{i}^{k}\}_{i=1}^{{N}_{f}}], \quad \dot{\mathbf{P}}_{i}^{k} =\{\dot{\mathbf{p}}_{i,1}^{k}, \dot{\mathbf{p}}_{i,2}^{k}, \cdots, \dot{\mathbf{p}}_{i,{N}_{p}}^{k}\},
\end{equation}
where $\dot{\mathbf{p}}_{0,0}^{k}$ represents the [\textit{CLS}] token of whole videos, $[\cdot]$ represents the concatenation operation. Note that $\mathcal{V}^{m} \in \mathbb{R}^{{N}_{f}*({N}_{p}+1)*d}$ while $\dot{\mathcal{V}}^{k}, \ddot{\mathcal{V}}^{k} \in \mathbb{R}^{({N}_{f}*{N}_{p}+1)*d}$.

The inputs of the first STAN layer $\dot{\mathcal{V}}^{1}$ are constructed from $\mathcal{V}^{m}$, where we first average the features of the [\textit{CLS}] token in each frame, obtaining $\dot{\mathbf{p}}_{0,0}^{1}=\frac{1}{{N}_{f}}\sum_{i=1}^{{N}_{f}}(\mathbf{p}_{i,0}^{m})$. We then update patch embeddings in $\dot{\mathcal{V}}^{1}$ with spatial and temporal position embeddings.

For the rest STAN layers, whose inputs $\dot{\mathcal{V}}^{k}$ are constructed based upon the outputs from the previous STAN layer $\ddot{\mathcal{V}}^{k-1}$ and visual encoder layer $\mathcal{V}^{m+k-1}$, which could be formulated as:
\begin{equation}
\label{eq:stan-rest}
	\left\{
	\begin{aligned}
		& \dot{\mathbf{p}}_{0,0}^{k} = \ddot{\mathbf{p}}_{0,0}^{k-1} + {\Theta}^{k}_{proj}(\frac{1}{{N}_{f}}\sum{\mathbf{p}_{i,0}^{m+k-1}}),
		\\
		& \dot{\mathbf{p}}_{i,j}^{k} = \ddot{\mathbf{p}}_{i,j}^{k-1} + {\Theta}^{k}_{proj}(\mathbf{p}_{i,j}^{m+k-1}),
	\end{aligned}
	\right.
\end{equation}
where $1 \leqq i \leqq {N}_{f}$, $1 \leqq j \leqq {N}_{p}$, and ${\Theta}^{k}_{proj} \in  {\mathbb{R}}^{d*d}$ denotes a linear projection layer.

Regarding the forward process of each layer, STAN first feeds the input features $\dot{\mathcal{V}}^{k}$ into a temporal self-attention module. To simplify notation, we omit the superscript of the inputs $\dot{\mathcal{V}}^{k}$ and represent the collection of $j^{th}$ patch embeddings in different frames as $\textbf{Y}_{j} \in \mathbb{R}^{{N}_{f}*d}$. In this way, the temporal propagation step at each specific spatial position could be formulated as: 
\begin{equation}
\label{eq:stan-temp}
	\dot{\textbf{Y}}_{j} = {\Theta}^{temp}_{proj} {\rm SA}({\rm LN}(\textbf{Y}_{j})),
\end{equation}
where ${\rm LN(\cdot)}$ represents the layer normalization operation, ${\Theta}^{temp}_{proj}(\cdot)$ is a temporal projection layer initialized as zero, ${\rm SA(\cdot)}$ denotes the self-attention computation, which could be formulated as:
\begin{equation}
\label{eq:stan-attn}
	\dot{\textbf{Y}} = {\rm softmax} \frac{\textbf{Y}{\Theta}_{q} {(\textbf{Y}{\Theta}_{k})}^{T} }{\sqrt{d}} (\textbf{Y}{\Theta}_{v}) + \textbf{Y}.
\end{equation}

STAN then exploits the multi-head self-attention mechanism within the visual encoder layer to construct its spatial self-attention module, whose parameters are also inherited during the initialization step. To simplify notation, we denote the input features of the $i^{th}$ frame as $\textbf{X}_{i} \in \mathbb{R}^{({N}_{p}+1)*d}$. In this way, the spatial propagation step at each temporal frame could be formulated as:
\begin{equation}
\label{eq:stan-temp}
	\dot{\textbf{X}}_{i} = {\rm SA}({\rm LN}(\textbf{X}_{i})).
\end{equation}

Ultimately, STAN combines the outputs of the last visual encoder layer $\mathcal{V}^{-1}$ and the last STAN layer $\ddot{\mathcal{V}}^{-1}$, obtaining the video features $\textbf{V}$, which could be formulated as: 
\begin{equation}
\label{eq:stan-temp}
	\textbf{V} = {\Theta}^{vis}_{proj}(\mathcal{V}^{-1} \oplus \ddot{\mathcal{V}}^{-1}),
\end{equation}
where ${\Theta}^{vis}_{proj}(\cdot)$ represents another linear projection layer to project video features into joint feature space. $\oplus$ denotes the addition calculation of the [\textit{CLS}] token features within $\mathcal{V}^{-1}$ and $\ddot{\mathcal{V}}^{-1}$.

Note that STAN is not our technical novelty, readers can refer to \cite{liu2023revisiting} and \cite{liu2023mug} for more details.

\subsection{Mug in Step 4}
\label{sec:supple-mug}

As illustrated in the top-right part of the overall framework in Figure \ref{fig:framework}, Mug attempts to filter out misaligned information among cross-modal features in an implicit manner. 

Specifically, Mug first derives a dot-product similarity matrix according to video features $\mathbf{V} \in {\mathbb{R}}^{{N}_{f}*d}$ and text features $\mathbf{T} \in {\mathbb{R}}^{{N}_{t}*d}$. Mug then assigns a frame-to-token attention
score ${z}_{i,j}$ to each text token features $\mathbf{t}_{j}$ based on its relevance to the current video frame features $\mathbf{v}_{i}$, whose calculation could be formulated as:
\begin{equation}
\label{eq:mug-t2v-1}
	{z}_{i,j} = \frac{ {\rm exp}(\tau \mathbf{t}_{j} \cdot \mathbf{v}_{i}) }{ \sum_{j=1}^{{N}_{t}} {\rm exp}(\tau \mathbf{t}_{j} \cdot \mathbf{v}_{i}) },
\end{equation}
where $\tau$ controls the sharpness of the attention distribution in Mug. We set $\tau=100$ in our experiments by default.

Afterward, for each frame, Mug aggregates initial text features based on the attention distribution, yielding frame-specific text features $\hat{\mathbf{t}}_{i} = \sum_{j=1}^{{N}_{t}}({z}_{i,j}\mathbf{t}_{j})$. Mug then measures the consistency of each frame features $\mathbf{v}_{i}$ with respect to the text tokens $\hat{\mathbf{t}}_{i} \in \mathbb{R}^{d}$ as:
\begin{equation}
\label{eq:mug-t2v-2}
	\hat{z}_{i} = \frac{ {\rm exp}(\tau \hat{\mathbf{t}}_{i} \cdot \mathbf{v}_{i}) }{ \sum_{n=1}^{{N}_{f}} {\rm exp}(\tau \hat{\mathbf{t}}_{n} \cdot \mathbf{v}_{n}) },
\end{equation}
where $\hat{z}_{i}$ represents the attention weight of each frame towards
the text. In this way, we aggregate all frame-wise features, obtaining the enhanced text-guided video features $\Bar{\Bar{\mathbf{v}}} = \sum_{i=1}^{{N}_{f}}(\hat{z}_{i}\mathbf{v}_{i})$.

Analogously, the process to generate enhanced video-guided text features $\Bar{\Bar{\mathbf{t}}}$ is a mirror operation as $\Bar{\Bar{\mathbf{v}}}$, whose calculation could be formulated as:
\begin{equation}
\label{eq:mug-v2t-1}
	{z}_{i,j}^{'} = \frac{ {\rm exp}(\tau \mathbf{t}_{j} \cdot \mathbf{v}_{i}) }{ \sum_{i=1}^{{N}_{f}} {\rm exp}(\tau \mathbf{t}_{j} \cdot \mathbf{v}_{i}) },
\end{equation}
\begin{equation}
\label{eq:mug-v2t-2}
	\hat{z}_{j}^{'} = \frac{ {\rm exp}(\tau \mathbf{t}_{j} \cdot \hat{\mathbf{v}}_{j}) }{ \sum_{n=1}^{{N}_{t}} {\rm exp}(\tau \mathbf{t}_{n} \cdot \hat{\mathbf{v}}_{n}) },
\end{equation}
where ${z}_{i,j}^{'}$ is the token-to-frame attention score, $\hat{\mathbf{v}}_{j} = \sum_{i=1}^{{N}_{f}}({z}_{i,j}^{'}\mathbf{v}_{i})$ is token-specific video features, and $\hat{z}_{j}^{'}$ represents the attention weight of each text token towards the video. In this way, we aggregate all token-wise features, obtaining the enhanced video-guided text features $\Bar{\Bar{\mathbf{t}}} = \sum_{j=1}^{{N}_{t}}(\hat{z}_{j}^{'}\mathbf{t}_{j})$.

Note that we exploit the enhanced video features $\Bar{\Bar{\mathbf{v}}} \in \mathbb{R}^{d}$ and text features $\Bar{\Bar{\mathbf{t}}} \in \mathbb{R}^{d}$ to compute the VTC loss $\mathcal{L}_{vtc}(\Bar{\Bar{\mathbf{v}}}, \Bar{\Bar{\mathbf{t}}})$, whose equation is the same as Eq. \ref{eq:VTC}. Mug is also not our technical novelty, readers can refer to \cite{liu2023mug} for more details.

\end{document}